\documentclass[nofootinbib,aps,pre,twocolumn,showpacs,superscriptaddress]{revtex4-2}

\usepackage{pratik}
\allowdisplaybreaks

\usepackage[color=gray!30,textsize=tiny]{todonotes}

\graphicspath{{./fig/}}

\makeatletter
\DeclareMathSizes{10}{9}{7}{5}
\makeatother

\begin{document}

\title{
An Analytical Characterization of Sloppiness in Neural Networks: Insights from Linear Models
}
\date{\today}

\author{Jialin Mao}\affiliation{University of Pennsylvania}
\author{Itay Griniasty}\affiliation{School of Mechanical Engineering, Tel Aviv University}
\author{Yan Sun}\affiliation{University of Pennsylvania}
\author{Mark K.\@ Transtrum}\affiliation{Brigham Young University}
\author{James P.\@ Sethna}\affiliation{Cornell University}
\author{Pratik Chaudhari}\affiliation{University of Pennsylvania}


\begin{abstract}
Recent experiments have shown that training trajectories of multiple deep neural networks with different architectures, optimization algorithms, hyper-parameter settings, and regularization methods evolve on a remarkably low-dimensional ``hyper-ribbon-like'' manifold in the space of probability distributions. Inspired by the similarities in the training trajectories of deep networks and linear networks, we analytically characterize this phenomenon for the latter. We show, using tools in dynamical systems theory, that the geometry of this low-dimensional manifold is controlled by (i) the decay rate of the eigenvalues of the input correlation matrix of the training data, (ii) the relative scale of the ground-truth output to the weights at the beginning of training, and (iii) the number of steps of gradient descent. By analytically computing and bounding the contributions of these quantities, we characterize phase boundaries of the region where hyper-ribbons are to be expected. We also extend our analysis to kernel machines and linear models that are trained with stochastic gradient descent.
\end{abstract}

\maketitle

\begin{figure*}[!htpb]
\centering
\includegraphics[width=0.335\linewidth]{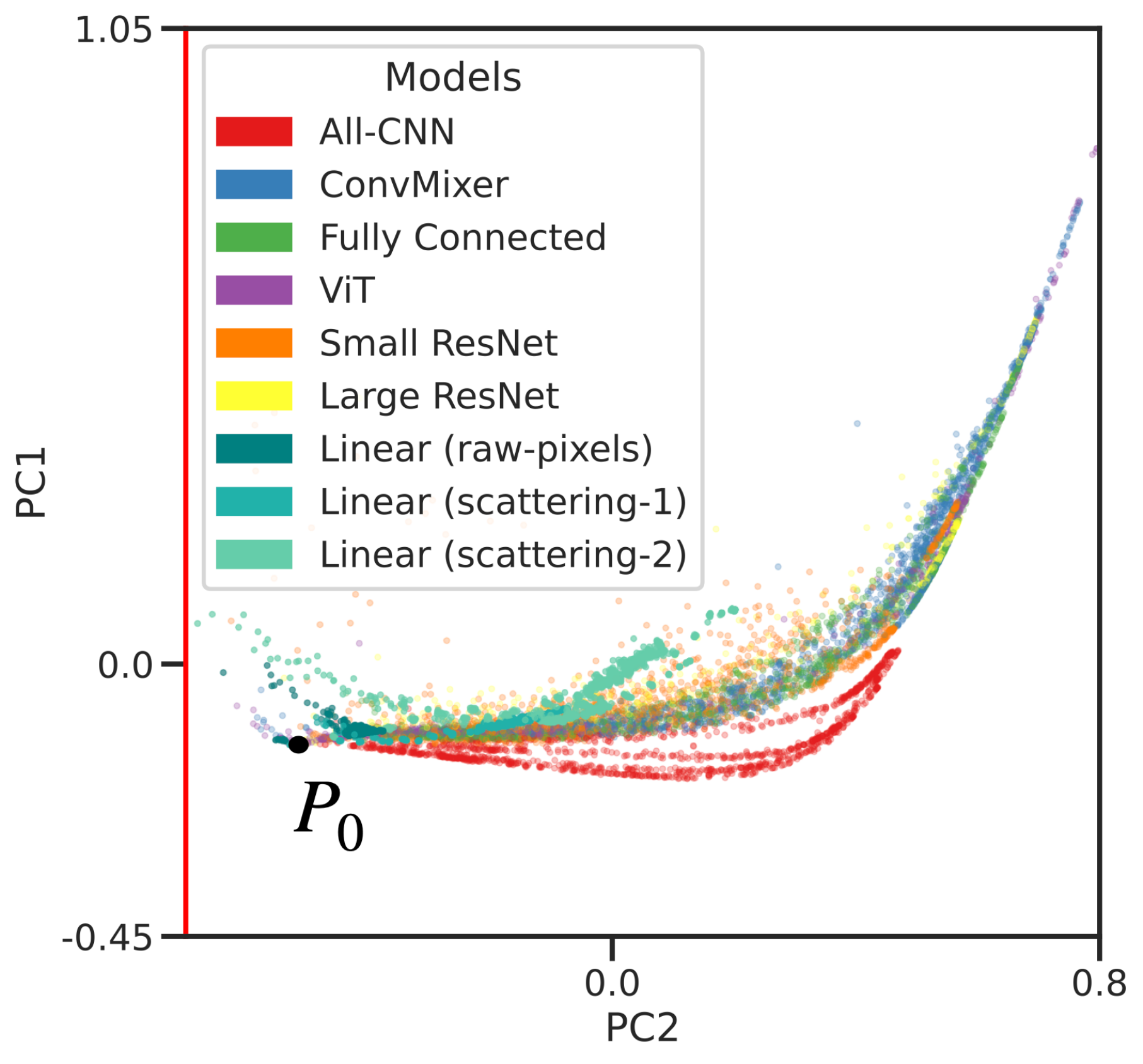}
\hspace*{2em}
\includegraphics[width=0.325\linewidth]{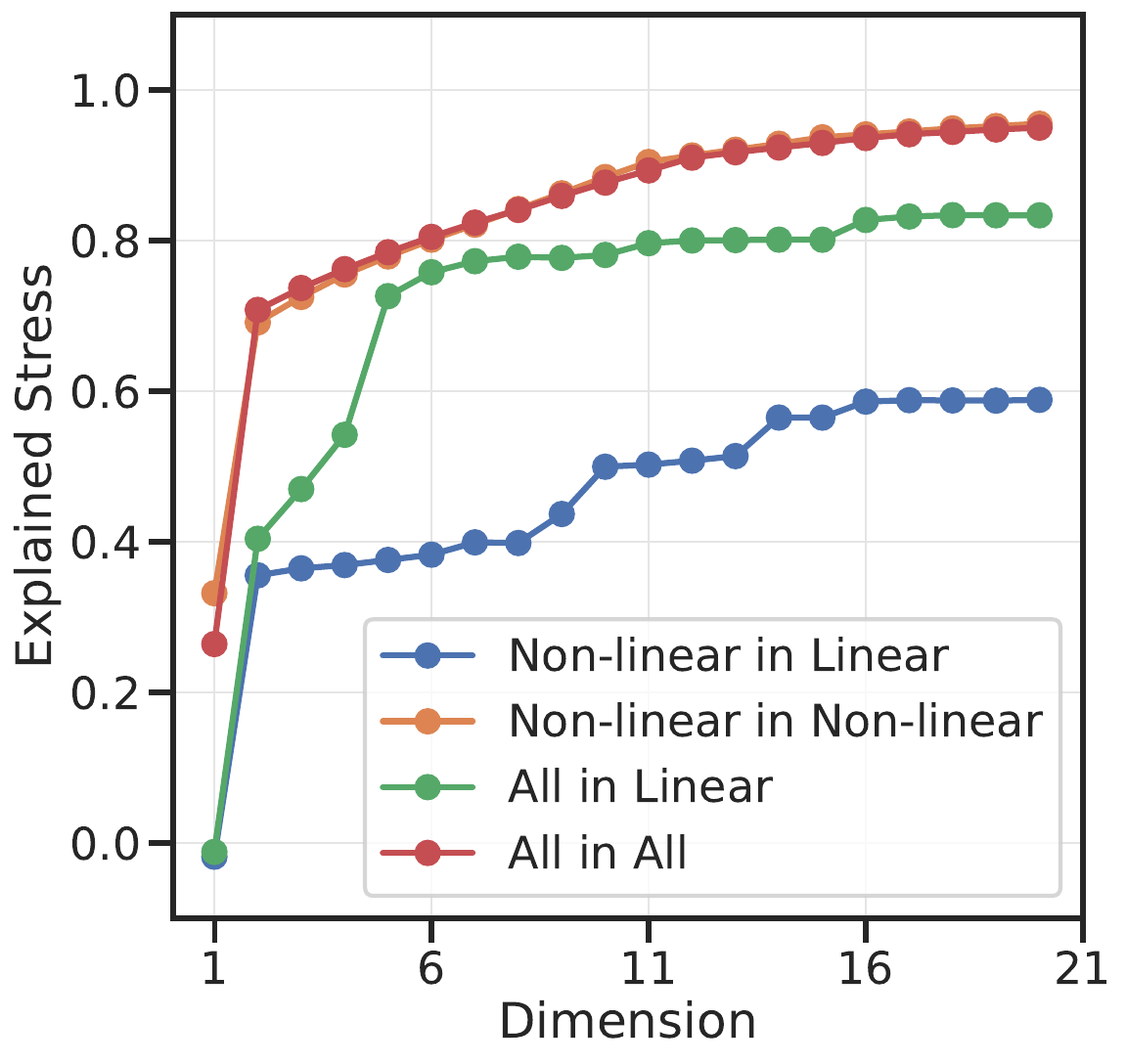}
\caption{\textbf{Left:}
The manifold of models along training trajectories of networks with different configurations (architectures denoted by different colors, optimization algorithms, hyper-parameters, and regularization mechanisms) is effectively low-dimensional for CIFAR-10. This is a partial reproduction using the data from \cite{mao2024training}. Linear networks are trained upon images directly (dark green), after pre-processing using one layer (``Scattering-1'' in lighter green) and two layers (``Scattering-2'' in the lightest green) of a scattering transform~\cite{mallat2012group}.
With typical weight initializations all models begin training near $P_0$, where every sample is assigned equal probability to belong to every class (marked by hand to guide the reader).
They progress to the truth $P_*$ (not seen here) to different degrees. All nonlinear deep networks in this experiment achieve zero training error. While linear networks do not fit the data perfectly, the manifolds swept by linear networks are quite similar to those of deep networks. These common low-dimensional manifolds in probability space are the inspiration of this paper.\\[0.25ex]
\textbf{Right:} A quantitative analysis of the inPCA embedding in terms of the explained stress which characterizes how well pairwise distances between points are preserved after the embedding. When the inPCA embedding is computed using all the points on the left, the explained stress of the first two dimensions is about 71\% (red), about the same when inPCA is computed using only nonlinear models (orange). We can compute inPCA using points corresponding to only linear models and embed all other models (green) or only the nonlinear models (blue) into this space. While there are clearly differences between the manifolds of linear and nonlinear models (blue curve is lower), it is remarkable that nonlinear models can be faithfully represented in the embedding constructed using linear models (green line is close to red and orange). Our analysis in this paper that focuses on linear models is therefore a meaningful insight into the manifolds of nonlinear models.
}
\label{fig:panel}
\end{figure*}

\section{Introduction}
\label{s:intro}

Recent experiments have shown that deep networks explore a low-dimensional manifold in the prediction space as they train to perform a task \cite{hacohen2020let,shamir2021dimpled,ramesh2022picture,mao2024training}. This manifold is dominated by its first few dimensions, with widths that decay geometrically. Remarkably, this manifold is shared between diverse networks, independent of their architecture, training algorithms and other specifications as depicted in \cref{fig:panel}. Low-dimensional structures also arise while fitting other non-linear models \cite{transtrumGeometryNonlinearLeast2011, quinn2019chebyshev}, where these phenomena have been explained by the fact that the function being used for approximation has a limited flexibility. But deep networks are universal approximators, and they are capable of fitting arbitrary datasets. There must be another cause for the low-dimensional manifolds in deep networks.

\begin{figure}[!htpb]
\centering
\includegraphics[width=0.65\linewidth]{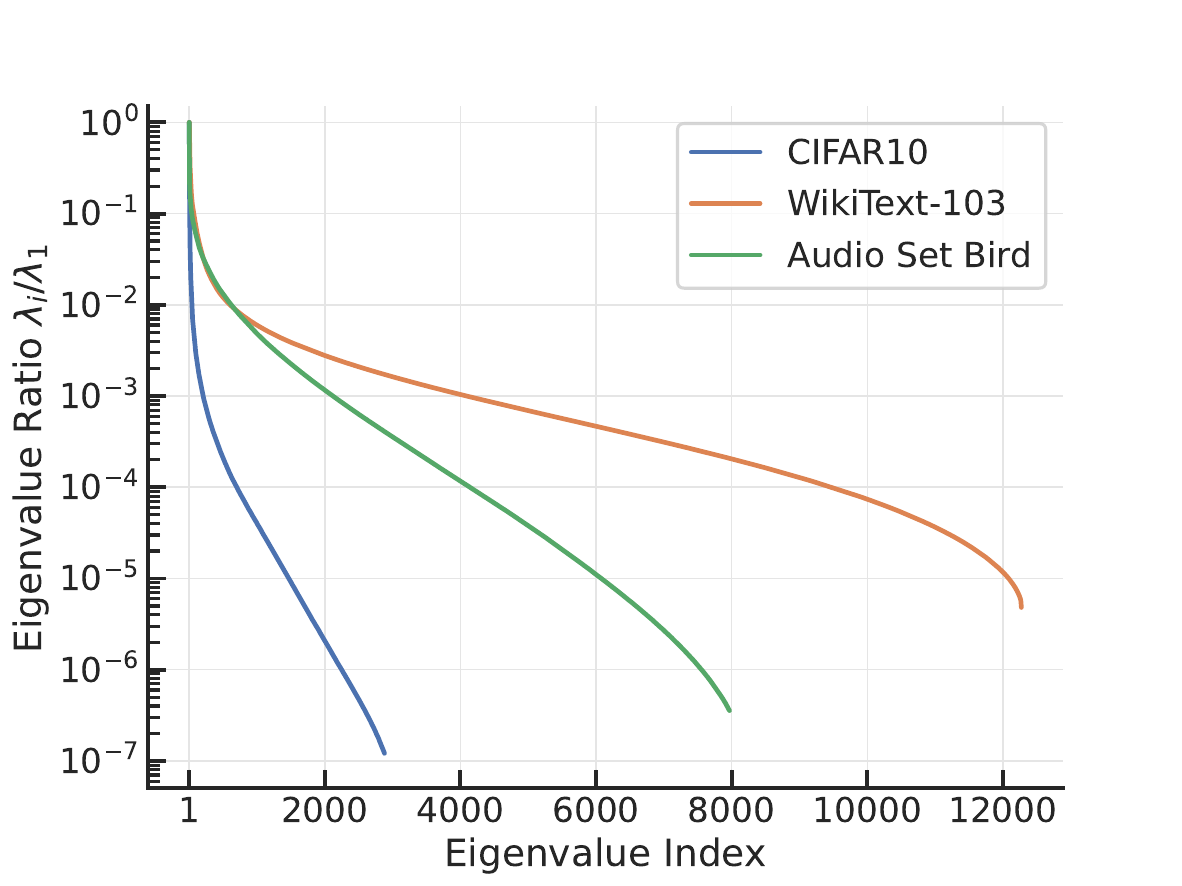}
\caption{\textbf{Real-world data are sloppy} Eigenvalues of the empirical correlation matrices of feature representations of different types of data exhibit a sharp drop-off among the first few eigenvectors. This is characteristic of multi-parameter models fit to data, but here it arises in real-world data. For CIFAR-10~\cite{krizhevsky2009learning}, we use raw pixel values as the features. For WikiText-103~\cite{merity2016pointer}, the text is broken into token sequences of length 16, and BERT~\cite{devlin2019bert} embeddings from the last hidden layer are concatenated to form a feature vector. For sounds corresponding to the Bird category in AudioSet \cite{gemmeke2017audio}, each 1-second audio segment is treated as a sample, with spectrograms serving as feature vectors.}
\label{fig:sloppy_eigen}
\end{figure}

In this paper, we argue that low-dimensional structures in training manifolds of deep networks arise not due to limited flexibility, but rather from the intrinsic low-dimensionality of the task. While non-linear models are ``sloppy", i.e., captured by manifolds in prediction space with geometrically decaying widths, due to their structural constraints, deep networks are sloppy because of the geometric properties of the task (examples in classical datasets are shown in \cref{fig:sloppy_eigen}). Inspired by the similarities in the training trajectories of deep and linear neural networks, we will exploit the analytical tractability of the latter to argue that it is indeed the task that results in the low-dimensionality of the training of deep networks. Let us first expand on the different facets of sloppiness.

\medskip \noindent \textbf{Model manifold}
Consider a dataset $\cbr{(x_i,y_i)}_{i=1}^n$ with inputs $x_i \in \reals^d$ and labels $y_i \in \{1,\dots, C\}$ which correspond to whether the corresponding input $x_i$ belongs to one out of $C$ different categories. A deep network-based classifier is a probabilistic model of this data. Given an input $x$, it assigns a probability to each possible output $y \in \{1,\dots,C\}$, this is denoted by $p_w(y \mid x)$ where $w \in \R^p$ are the parameters/weights of the model. If $\vec y = (y_1,\dots, y_n) \in \{1,\dots,C\}^n$ is the set of all possible outputs for a dataset, we may write
\(
    P_w(\vec y) = \prod_{i=1}^n p_w(y_i \mid x_i)
\)
as the joint probability assigned to these outputs by the model. As we vary the weights $w$ over some region in $\R^p$, the probabilistic model $P_w$ spans a region that is a subset of the $n (C-1)$-dimensional simplex. Let us call this set the ``model manifold''. Statistical divergences, such as the Kullback–Leibler (KL) divergence, quantify distances between probabilistic models and thereby induce a natural geometry on the weight space. The Fisher Information Matrix (FIM) is the natural Riemannian metric on the model manifold~\cite{amariInformationGeometryIts2016}.

\medskip \noindent \textbf{Training manifold}
Now consider \cref{fig:panel} again, which was partially reproduced from~\cite{mao2024training}. The authors developed techniques to analyze the model manifold of deep networks. They showed that probabilistic models corresponding to different points on the training trajectories of multiple deep networks with different architectures, optimization algorithms, hyper-parameter settings, and regularization methods evolve on a remarkably low-dimensional model manifold. To visualize and analyze this manifold, they used a technique known as intensive principal component analysis (inPCA~\cite{quinn2019visualizing}), closely related to multi-dimensional scaling \cite{cox2008multidimensional} and PCA~\cite{hotelling1933analysis}, to embed the models into a Minkowski space. While an exact isometric embedding is guaranteed when the dimensionality of the embedding space is equal to the number of models, they found that low-dimensional projections preserved pairwise distances between probabilistic models (statistical divergences such as the Bhattacharyya distance) very well. A two-dimensional embedding shown in~\cref{fig:panel} captures 71\% of the ``stress'' (i.e., pairwise distance preservation). A mere 50 dimensions are sufficient to account for 98\% of the stress, even when these architectures range from 0.6 million to 50 million weights. They observed that the stress explained by each successive dimension decays geometrically, indicating a highly concentrated spectrum. The points corresponding to one particular architecture in~\cref{fig:panel}, which is the set of models explored during the training process, are a subset of the model manifold of that architecture. It is evident that this subset is very low-dimensional. Let us call it the ``training manifold''.

\medskip \noindent \textbf{Sloppy models}
The geometric decay in the stress explained by each dimension is strikingly reminiscent of structural patterns found in another class of over-parameterized models known as sloppy models~\cite{quinnInformationGeometryMultiparameter2021}. Sloppiness was first identified in systems biology, where detailed mechanistic models---large systems of differential equations with numerous unknown rate constants---were constructed to describe protein or gene regulation networks~\cite{brown2003signal}. In practice, inferring these parameters from data is nearly impossible. But one need not infer them exactly, analyses based on the Fisher Information Matrix and the Cramér–Rao bound~\cite{keener2010theoretical} showed that these parameters could vary by many orders of magnitude without substantially affecting the fit to experimental data, or predictions on new data. The signature of such ``sloppy'' models is that successive eigenvalues of the FIM computed at the fitted parameters decay geometrically. Sloppy models consistently exhibit good generalization---not only to out-of-sample data, but often to out-of-distribution conditions, such as predictions under novel drug regimens that differ significantly from those used for model calibration~\cite{transtrumModelReductionManifold2014a}.

\medskip \noindent \textbf{Sloppy data}
Deep networks exhibit a lot of these same phenomena, e.g., an overwhelmingly large number of eigenvalues of the Hessian and FIM of typical networks are vanishingly small \cite{chaudhari2016entropy,papyan2018full,chaudhari2017stochastic}. Symmetries in multi-layer architectures entail that both these matrices have a large number of zero eigenvalues \cite{pmlr-v70-dinh17b,sunLightlikeNeuromanifoldsOccam2020a,watanabe2007almost}. The authors in \cite{yang2021does} showed that the eigenspectrum of the input correlation matrix is closely related to the eigenspectrum of the Hessian of the training loss and the FIM for general deep networks. They used this observation to argue that deep networks generalize---in spite having fewer data than the number of parameters---when the input correlation matrix has eigenvalues that decay geometrically. Let us call such data ``sloppy''. Almost pervasively, real-world data exhibits this sloppy structure \cite{ramesh2024many}, for example \cref{fig:sloppy_eigen} shows that eigenvalues of the input correlation matrices of three different modalities (images, text, and sounds) span a very large range and are spread nearly uniformly on a logarithmic scale. This structure in the data has been argued to be the principle underlying effective continuum and universal theories in physics \cite{machtaParameterSpaceCompression2013}.

\medskip \noindent \textbf{Hyper-ribbons}
The FIM can exhibit geometric decay in its eigenvalues even when input data is one dimensional, i.e., when it could not be sloppy. Consider the situation when the targets are given by $y = f(x ,w) \in \R$ for $x \in [-1,1]$ and weights $w \in \R^p$. The truncated Taylor series of $f$ may be written as
\[
    f(x, w) \approx f_n(x, w) = \sum_{k=0}^{n-1} \varphi_k(w) x^k
\]
where $\varphi_k(w) = \f{f^{(k)}(\xi, w)}{k!}$ for $\xi \in [-1,1]$ and $f^{(k)}$ denotes the $k^{\text{th}}$-derivative of $f$ with respect to the data $x$. If the true function has a radius of convergence $R > 1$ and if $w$ varies over some compact region, say, $\sum_k w_k^2 \leq 1$, then
\[
    \abs{\varphi_k(w)} \leq c R^{-k}
\]
for some constant $c$. The model manifold of $f$ is contained within a distance $\OO(R^{-n+1})$ of the model manifold of $f_n$. If we consider a dataset with $n$ samples, the model manifold of $f_n$, which is the set of all possible predictions on these samples, is a subset of $\R^n$. In fact, it is a hyper-ellipsoid where the lengths of the principal axes decay geometrically. This is because the Jacobian of the predictions in $\R^n$ with respect to the weight space is a Vandermonde matrix which has geometrically decaying singular values~\cite{waterfallSloppyModelUniversalityClass2006,waterfall2006universality}. The FIM, which is the outer-product of this Jacobian, inherits this structure, its eigenvalues also decay geometrically. The FIM is a purely local property, but it is closely related to the global geometry of model manifold, in our example. Such model manifolds look like ``hyper-ribbons''. There is one set of parameters that is tightly constrained by the data, another which can vary, say, twice as much without affecting predictions, and so on. This hyper-ribbon structure reflects the intrinsic limitations in model flexibility: the model simply cannot produce a wide range of predictions, despite its many parameters.

\subsection{Contributions of this manuscript}

These four concepts are clearly distinct from each other, but they share a suspiciously common pattern: (i) the stress captured by successive dimensions of an inPCA embedding of the training manifold of deep networks, (ii) eigenvalues of the FIM of sloppy models at typical fits, (iii) eigenvalues of the input correlation matrices of real-world data across diverse modalities, and (iv) cross-sectional widths of hyper-ribbon-like model manifolds---all exhibit geometric decay. The goal of this manuscript is to identify and clarify connections between these concepts.

The key question that will help frame our narrative is: why do different architectures, training and regularization methods, explore such a tiny subset of the prediction space? Deep networks are universal approximators~\cite{cybenko1989approximation}. With sufficient capacity and training, they can be induced to fit even anomalous or completely random data. The model manifold of a deep network-based classifier is therefore high-dimensional, it spans the entire $n (C-1)$-dimensional simplex. However, across large problems with $n (C-1) \sim 10^6$--$10^8$, the authors in~\cite{mao2024training,rameshPictureSpaceTypical2023} have found training manifolds to be remarkably low-dimensional, as low as 50 dimensions seem sufficient to embed the manifold faithfully. The same story holds for regression. This flexibility implies that the manifolds explored by deep networks during training cannot be hyper-ribbons in the same sense as those in sloppy models. While they may exhibit low-dimensional structure empirically, this structure cannot arise from a fundamental limitation in model flexibility. It requires a different theoretical explanation.

We show in \cref{fig:panel} that the training manifold of linear models, e.g., linear predictors fitted upon raw pixels, or after projecting input images upon a fixed nonlinear basis computed using the scattering transform~\cite{mallat2012group}, is similar to that of deep networks. This empirical observation motivates our analysis.
\begin{figure}[!htpb]
\centering
\includegraphics[width=0.65\linewidth]{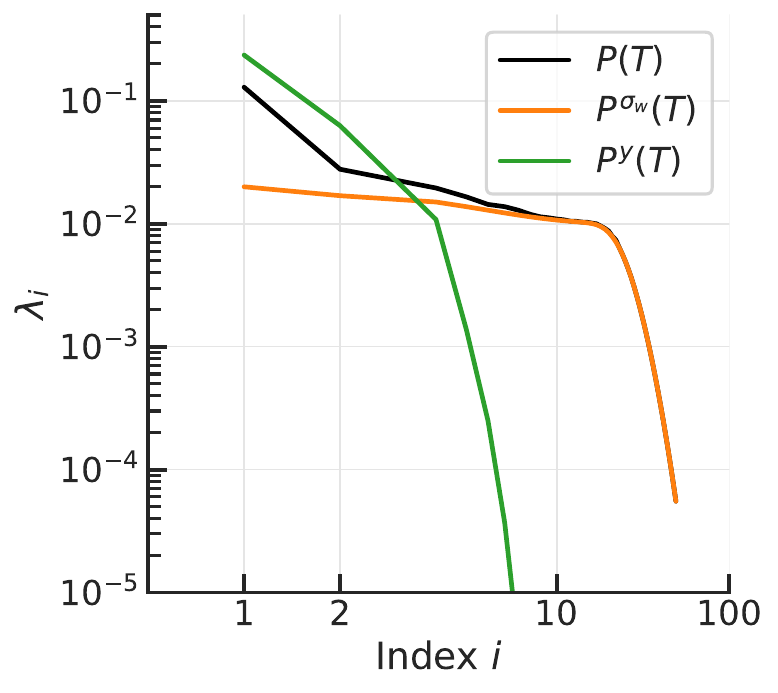}
\caption{
\textbf{Eigenvalues of PCA of points on training trajectories of linear regression and eigenvalues of different important contributions to it.}
There is a rapid decrease in the PCA eigenvalues (black), and thus these points admit a low-dimensional representation with high explained-variance in the first few dimensions, similar to linear and nonlinear models in \cref{fig:panel}. The PCA covariance matrix $P(T)$ after $T$ gradient descent steps can be decomposed analytically into two important contributions: $P^y(T)$ that depends upon the regression targets and $P^{\sigma_w}(T)$ that depends upon the distribution of initial weights. The sharp initial decay in the eigenvalues of $P(T)$ is well-approximated by $P^y$ while the decay in the tail is controlled by $P^{\sigma_w}$. Details in~\cref{fig:decomposition}.
  }
\label{fig:decomposition_intro}
\end{figure}
\cref{fig:decomposition_intro} encapsulates some of the main results of this manuscript, discussed in~\cref{s:linear_regression}. We analyze how and when low-dimensional representations of high-dimensional training trajectories arise for linear regression problems (as we have found for the real-world models in \cref{fig:panel}). The key players of this analysis and their predominant roles are as follows.
\begin{itemize}[nosep]

\item The sloppiness in the input data being fit, characterized by the logarithm of the ratio of successive eigenvalues of the input correlation matrix, we will call this the
\[
    \text{slope: } c,
\]
that mimics the eigenvalue decay in ~\cref{fig:sloppy_eigen}. Sloppy data allows the first few dimensions of the hyper-ribbon of training trajectories to capture most of its variance.

\item The relative scale of the ground-truth weights to the initial weights which are drawn from zero-mean Gaussian distributions with isotropic variance $\s_*^2$ and $\s_w^2$ respectively,
\[
    \text{weight initialization strength: } \f{\s_*}{\s_w}.
\]
The spread of initial conditions leads to a spread in trajectories that will dominate the tail of the eigenspectrum.

\item The
\[
    \text{number of weight updates: } T.
\]
Larger the $T$, larger the volume of the output space explored by the different training trajectories and therefore, intuitively, larger the dimensionality of the hyper-ribbon.
\end{itemize}

In the sequel, we will make more precise statements that relate these three parameters to training trajectories.
Our analysis will proceed by analyzing the principal components analysis (PCA) matrix $P(T)$ of points along training trajectories obtained by initializing linear models at different points, and fitting them upon data with different degrees of sloppiness (different $c$).
We will be able to decompose $P(T)$ into three pieces, the two most important being a piece $P_1^y$ determined predominantly by the ground-truth targets, and a piece $P_1^{\sigma_w}$ determined by initialization, the relative importance of these terms is proportional to the weight initialization strength. By analytically computing and bounding these various contributions (\cref{fig:decomposition_intro}), we characterize the ``phase boundaries'' of the region where low-dimensional hyper-ribbons are to be expected (\cref{fig:contour_plots}). In~\cref{s:modifications} we will also extend our analysis to kernel machines and linear models that are trained with stochastic gradient descent.

\section{Training manifold for linear regression}
\label{s:linear_regression}

\subsection{Target data set, training, and weights}
Consider a dataset $\cbr{(x'_i,y_i)}_{i=1}^n$ that consists of inputs $x'_i \in \reals^{d-1}$ and outputs $y_i \in \reals$. We will focus on the case with a scalar output in this paper for clarity of exposition, all results hold for multi-dimensional output. Let $x_i \equiv [x'_i, 1]$ denote the input with a constant appended to it and consider a linear model $y_i = w^\top x_i$ with $w \in \reals^d$ trained to minimize

\beq{
    C(w) = \f{1}{2n} \sum_{i=1}^n r_i(w)^2.
    \label{eq:C}
}
Here the residuals $r_i(w) = \hat y_i - y_i$ for $i \in \cbr{1,\dots,n}$ denote the difference between the predictions and the targets \(y_i\in \reals\). We will assume that targets correspond to unknown true weights $w^* \in \reals^d$, i.e., $y_i = {w^*}^\top x_i$. Discrete-time gradient descent to minimize this objective with a step-size $\a$ (learning rate) can be written as
\(
w_{t+1} = w_t - \a\ \partial_w C(w_t),
\)
starting from some $w_0 \in \R^d$ for all $t=1,2,\dots$.
We denote the $n$-dimensional vector of residuals computed at weights $w_t$ by $r_t \equiv [r_1(w_t), \dots, r_n(w_t)]^\top \in \R^n$. As the weights are updated by gradient descent, this vector $r_t$ evolves as
\beq{
r_{t+1} = (I - \a K) r_t = (I - \a K)^{t+1} r_0,
\label{eq:rtp}
}
where $I \in \R^{n\times n}$ is the identity matrix and $K  \in \R^{n\times n}$ is a symmetric positive semi-definite matrix with entries $K_{ij} = x_i^\top x_j/n$ for $i,j \in \{1,\ldots,n\}$. Note that $K = X X^\top/n$ where the $i^{\text{th}}$ row of $X \in \R^{n \times d}$ contains the input datum $x_i$. In other words, for linear models, the neural tangent kernel (NTK~\citep{jacot2018neural}) is simply the input-correlation matrix. Let the $i^{\text{th}}$ largest eigenvalue of $K$ be denoted by $\l_i^K \geq 0$ . The shorthand
\[
     K_d \equiv I - \a K,
\]
will be useful to simplify our expressions. It is the first-order approximation of $\exp(-\a K)$.
We make the following assumptions in our analysis.
\begin{enumerate}[(i),nosep]
\item \textbf{Input data and ground-truth targets:} We will assume that input data are sloppy with a decay rate $c > 0$ on the logarithmic scale, i.e.,
\[
\l_i^K = \exp(-(i-1) c),
\]
for all $i=1, \dots, n$, with \( c\gg 1/n \).
We will assume that the unknown true weights $w^*$ are a random variable
\[
\R^d \ni w^* \sim N(0, \s_*^2 I)
\]
where $\s_*^2$ is a scalar. This will indirectly be an assumption on the norm of the targets $y_i$.
\item \textbf{Model:} The model is in the data scarce regime, i.e., the number of samples is smaller than the dimensionality of the input data $n < d$. Weights $w$ are under-determined if $n < d$ and therefore an infinite set of solutions achieves $C(w) = 0$. We will assume that $\rank(K) = n$. This is not unduly restrictive. All our analysis can also be conducted in the image space of $K$ if $K$ is rank deficient.

\item \textbf{Training method:} For a large part of the analysis we will be interested in training methods that resemble gradient descent. We will assume that the step-size $\a<1/\l_1^K$. This assumption ensures that $\norm{I-\a K}_2 < 1$ and therefore $\|r_t\|_2 \to 0$ monotonically as $t \to \infty$ and therefore it is quite standard in the analysis of gradient descent algorithms~\citep{bottou2012stochastic}. The most important parameter of the training process for us will be the total number of weight updates $T$.

\item \textbf{Weight initialization:}
We will assume that weights are initialized to values $w_0$ sampled from a Gaussian distribution with zero mean and an isotropic variance
\[
\R^d \ni w_0 \sim N(0, \s_w^2 I).
\]
for a scalar $\s_w^2$. The ratio $\s_*^2/\s_w^2$ controls the distance in weight space that a training trajectory needs to travel from its initialization at $w_0$ to fit to the true weights $w^*$. We will see that, roughly speaking, larger this ratio, smaller the dimensionality of the hyper-ribbon and weaker the dependence on the other parameters of the problem such as the slope $c$ or the number of weight updates $T$.
\end{enumerate}

\subsection{Principal component analysis of the training manifold}
Consider $N$ randomly initialized models with weights
$\{w_0^{(i)}\}_{i=1}^N$. From assumption (iv) on weight initialization, we have $\E[w_0] = 0$ and $\E[w_0 w_0^\top] = \s_w^2 I$. The initial residual vectors satisfy
\[
\E \sbr{r_0^{(i)}} = -y, \quad \E \sbr{r_0^{(i)} {r_0^{(j)}}^{\top}} =  n \d_{ij} \s_w^2 K + yy^{\top},
\]
for all $i, j \in \cbr{1,\dots, N}$ where $y = [y_1, \dots, y_n]^\top\in \R^n$ is the vector of targets and $\d_{ij} = \mathbf{1}_{\{i=j\}}$ is the delta function. The prediction space $\R^n$ has Euclidean geometry. We can therefore capture the geometry of the training manifold, which is a subset of  $\R^n$, using principal component analysis (PCA) of points along trajectories $\{r_t^{(i)}\}_{t=0,\,i=1}^{T-1,\,N}$. The covariance matrix corresponding to PCA is
\begin{equation}
\begin{split}
\label{eq:rpca_finite}
    P(N, T)
    &= \frac{1}{NT} \sum_{i=1}^N \sum_{t=0}^{T-1} (r_t^{(i)} - \bar{r}) (r_t^{(i)} - \bar{r})^\top,\\
    &= \frac{1}{N} \rbr{\sum_{i = 1}^{N} \frac{1}{T}\sum_{t=0}^{T-1} r_t^{(i)} {r_t^{(i)}}^\top} - \bar r\ \bar r^\top,
\end{split}
\end{equation}
 where the mean residual is
\(
    \bar r \equiv \bar r(N,T) = \frac{1}{N T} \sum_{i, t} r_t^{(i)}
\)
for $T \geq 1$.
The mean residual evolves according to the equation
\(
\bar r (N, T)= K_T  \bar r(N, 1)
\)

with
\beq{
    K_T \equiv \frac{1}{T} \sum_{t = 0}^{T-1} K_d^t= \frac{1}{\a T} K^{-1} (I- K_d^T).
    \label{eq:K_T}
}
As the number of random initializations $N$ goes to infinity, we can separate the PCA matrix $P$ into two components,
\beq{
    \aed{
    P(T) &= \lim_{N \to \infty} P(N, T) \\
    &= \frac{1}{T}\sum_{t = 0}^{T-1} K_d^t (\s_w^2 K + y y^{\top}) K_d^t - \underbrace{K_T y y^\top K_T}_{\equiv P_2(T)}\\
    &\equiv \underbrace{P_1^ {\s_w}(T)+P_1^ {y}(T)}_{\equiv P_1(T)} - P_2(T).
    }
   \label{eq:P}
}
We have broken down the PCA matrix into its three components.
The third term $P_2(T)$ has unit rank. The first term $P_1^ {\s_w}(T)=\frac{n\s_w^2}{T}\sum_{t = 0}^{T-1} K_d^{2t}  K$ depends on the variance of initial weights $\s^2_w$, and the second term $P_1^{y}(T)=\frac{1}{T}\sum_{t = 0}^{T-1} K_d^t  y y^{\top} K_d^t$ again depends on the targets $y$, but it is not unit rank. 
We will analyze this expression further in the next section.

\subsection{The geometry of the training manifold}

The goal of this section will be to characterize the geometry of the training manifold, i.e., eigenvalues of the PCA matrix $P(T)$ in~\cref{eq:P}.
To simplify the notation, we will denote eigenvalues of matrices $P_2$, $T P_1^y$ and $T P_1^{\s_w}$ by \(\l^{P_2}\), \(\l^{\s_w}_i\) and \(\l^y_i\), respectively, for all $i = 1, \dots, n$. The scaling with $T$ on the latter two matrices is only for the sake of convenience in our exposition.

\medskip \noindent \textbf{Eigenvalue of $P_2$.}
For any two real symmetric matrices $A,B \in \R^{n \times n}$, Weyl's inequality says that
\beq{
    \l_i^A + \l_n^B \leq \l_{i}^{A+B} \leq \l_i^A + \l_1^B,
    \label{eq:weyl}
}
for eigenvalues $\l_i$ ordered in decreasing order. The third term $P_2(T)$ in~\cref{eq:P} is an outer product of $y$ with itself and has a single non-vanishing eigenvalue $\l^{P_2} = \norm{K_T y}^2$. Therefore, eigenvalues of $P(T)$ are sandwiched by the eigenvalues of $P_1(T)$:
\beq{
    \aed{
    \max \rbr{\l_{i+1}^{P_1}, \l_{i}^{P_1} - \l^{P_2}}  \leq \l_i^P \leq \l_i^{P_1}.
    }
    \label{eq:weyls_P}
}
From~\cref{eq:K_T},
\beq{
    \norm{K_T}_2
    \leq \f{1- (1-\a \l_1^K)^T}{\a T \l_n^K}
    \leq \f{1}{\a T \l_n^K}.
    \label{eq:P2}
}
This suggests that $\l^{P_2} = \mathcal{O}(1/T^2)$. The approximation becomes tighter for long training times. We can therefore focus our analysis on understanding $P_1(T)$.

\medskip \noindent \textbf{Weight initialization strength contribution.}
Since $K_d = I - \a K$ commutes with $K$, we can write $T P_1^ {\s_w}(T)$ to be the geometric sum
\[
   { T P_1^ {\s_w}(T) = n \s_w^2 \sum_{t = 0}^{T-1} K_d^{2t} K = \frac{n\s_w^2}{\a} (I+K_d)^{-1} (I - K_d^{2T}),}
\]
and calculate its eigenvalues explicitly as
\begin{equation}
{\l_i^{\s_w} = \f{n \s_w^2}{\a} \rbr{\frac{1 - (1 - \a \l_i^K)^{2T}}{2 - \a \l_i^K}}}.
\label{eq:lambda_sw}
\end{equation}
Due to the inequality $1 - (1-x)^a \leq \min\{1, ax\}$, which holds for $\abs{x} < 1$ and $a \geq 1$, under assumption~(iii), the eigenvalue $\l_i^{\s_w}$ is bounded above by
\begin{equation}
    \label{eq:w1_eval_bound}
    {\l_i^{\s_w} \leq \f{n\s_w^2}{\a} \rbr{\frac{\min\{1, 2T \a \l_i^K\}}{2 - \a \l_i^K}}}.
\end{equation}
\begin{figure}[!htpb]
\centering
\includegraphics[width=0.8\linewidth]{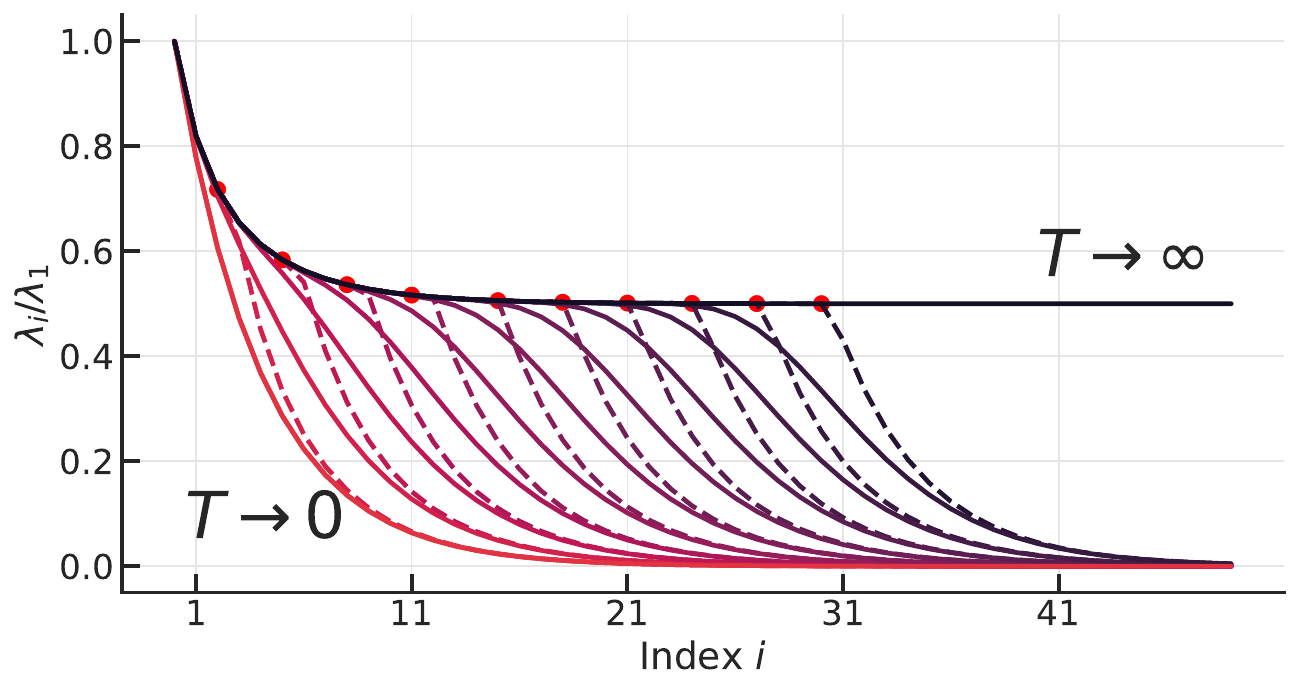}
\caption{
\textbf{Contributions to the eigenspectrum of the PCA matrix coming from weight initialization computed using the bound in~\cref{eq:w1_eval_bound} (dotted) and numerical computation of the corresponding term in~\cref{eq:P} (bold) with $\s_w^2 = 1$, $\a = 1$ and $c  = 0.5$.}
For very large training times $T \gg \l_1^K / \l_n^K$, the eigenvalue $\l_i^{\s_w}$ corresponds to the minimum in the numerator being 1 for any $i$. This means that $\l_i^{\s_w}$ decays to $\sim \s_w^2/2\a$ as the index $i$ increases, at a rate determined by the decay of $\l_i^K$.
From~\cref{eq:w1_eval_bound}, for $T \ll \l_1^K / \l_n^K$ and $i < \ln(2T\a) / c$, the minimum in the numerator is 1, and we again have a similar decay. For larger values of $i$, the minimum comes from the other term and therefore $\l_i^{\s_w}$ decays to much smaller values $\sim \f{1}{(2/\l_n^K - \a)}$. This is the lower envelope of the curves above.
The limit $T \to 0$ corresponds to the eigenvalues of $\s_w^2 K$ and therefore reflects the sloppy decay in the input correlation matrix.
}
\label{fig:w1_eval_bound}
\end{figure}
\cref{fig:w1_eval_bound} uses~\cref{eq:w1_eval_bound} to explain how eigenvalues $\l_i^{\s_w}$ of different indices $i$ depend upon the training duration $T$ and the sloppy eigenspectrum of the input correlation matrix $K$.
From~\cref{eq:P}, it is immediate that if the initialization variance $\s_w^2$ is small (relative to $\s^2_*$ which controls $\norm{y}$), the contribution of $P_1^{\s_w}(T)$ to the dimensionality of the hyper-ribbon is small for all times $T$.
Our calculation shows that for all times $T$, the head of the eigenspectrum $P_1^{\s_w}$ decays rather quickly.
For small times $T$, eigenvalues in the tail of $P_1^{\s_w}$ are quite small.
The implication is that, everything else (i.e., $P_1^y$) being the same, models trained for long times have a higher-dimensional hyper-ribbon due to variations caused by the initialization of weights. For short times, the hyper-ribbon has a smaller dimensionality.

\medskip \noindent \textbf{Target contribution.}
The second term corresponding to $P_1^y(T)$ in~\cref{eq:P} resembles the so-called reachability Gramian in systems theory. It is well-known that $P = \lim_{T\to \infty} P_1^y(T)$ is the unique solution to the discrete algebraic Lyapunov equation~\citep{numericalSolutionsLyapunov}
\[
    K_d P K_d^\top - P + y y^\top = 0.
\]
In systems theory, this concept is used for model reduction, i.e., to identify a low-dimensional dynamical system that captures time-varying data from a larger system. The rate of decay of the singular values of the reachability Gramian characterizes the quality of this approximation. Singular values of the Gramian decay quickly~\cite{penzlEigenvalueDecayBounds2000,Antoulas_2005,townsendSingularValuesMatrices2018a,Beckermann_Townsend_2016}
when $K_d$ has some nice properties (e.g., normal, well-conditioned), and $y y^\top$ is approximately low rank. This is precisely the setting of our paper.
To study $P_1^y(T)$, which is a finite sum, we write it as the difference between two Gramians:
\[
\aed{
T P_1^y(T)
&= \sum_{t = 0}^{\infty} K_d^t yy^\top K_d^t - \sum_{t=0}^\infty K_d^t \rbr{K_d^T yy^\top K_d^T} K_d^t \\
&= \sum_{t = 0}^{\infty} K_d^t \rbr{yy^\top - K_d^T yy^\top K_d^T} K_d^t.
}
\]
We summarize our results in 3 lemmas whose proofs are given in the \cref{app:proofs}.
The following lemma shows that the eigenspectrum of $P_1^y(T)$ decays quickly.
\begin{lemma}
\label{lemma:w0_decay_bound}
We have
\[
\frac{\l_{1 + 2i}^y}{\l_1^y}
\leq \frac{4\rho^{-2i}}{(1+\rho^{-4i})^2} <
4\rho^{-2i},
\]
where
\[
\rho =  \exp\left( \frac{\pi^2}{2\ln\bigl(8\l_1^K/\l_n^K-4\bigr)} \right).
\]
\end{lemma}
The following lemma now gives a lower bound on the eigenvalue $\l_1^{y}$.
\begin{lemma}
\label{lemma:lambda_y_bound}
If we denote  $\tilde{\l}_i = \sum_{t = 0}^{T-1} (1 - \a \l_i^K)^{2t}$, then
\[
 \norm{y}^2 \leq \l_1^y\leq \tilde{\lambda}_n \norm{y}^2,
\]
where the norm of the targets concentrates around the value
\beq{
    { \norm{y}^2 \approx n \s_{*}^2\sum_{i=1}^n \l_i^K.}
    \label{eq:norm_y}
}
Notice that $\tilde{\lambda}_i  = \l_i^{\s_w}/(\s_w^2 \l_i^K)$.
\end{lemma}

\cref{lemma:w0_decay_bound} suggests that the eigenspectrum of $P_1^y$ decays as $\exp(-i \pi^2/ n c)$. In contrast to the decay of $P_1^{\s_w}$, this rate is independent of the training time $T$.
Suppose now that $\s_w^2$ is small relative to $\s_*^2$. Since $\l_1^y \geq \norm{y}^2 =n \s_*^2 \tr(K)$, the head of the eigenspectrum of $P_1^y$ can be much taller than that of $P_1^{\s_w}$, even if the decay of the two is similar. In other words, the head of eigenspectrum of $P_1(T) = P_1^{\s_w}(T) + P_1^y(T)$ is determined by $P_1^y$ and the tail is determined by $P_1^{\s_w}$.

\begin{figure}[!htpb]
\centering
\includegraphics[width=0.65\linewidth]{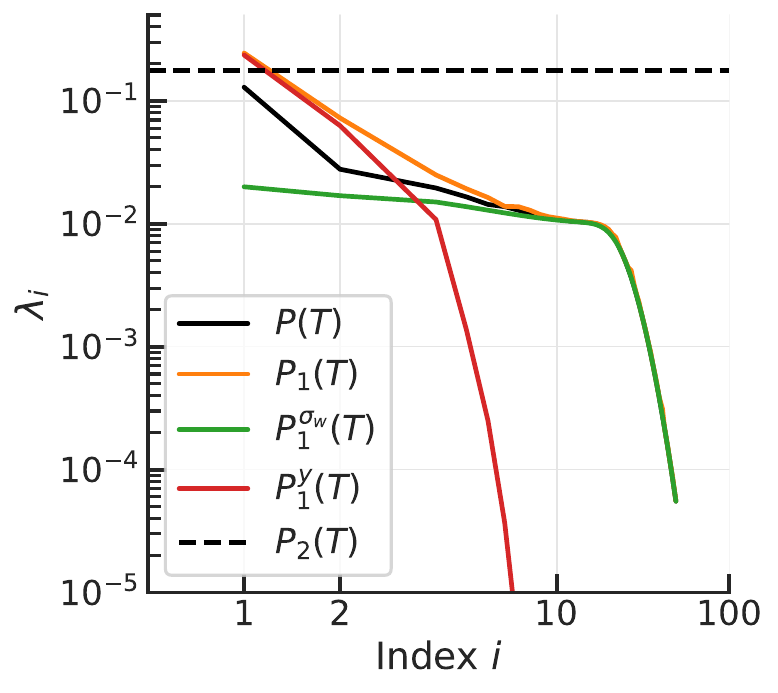}
\caption{
\textbf{The tail of the eigenspectrum of $P_1(T)$ is well-approximated by the contribution coming from the weight initializations $P_1^{\s_w}(T)$, while the head is well approximated by the contribution coming from the targets $P_1^y(T)$.}
These eigenvalues were computed for $d=100$ dimensional data with slope $c=0.2$ for the eigenvalues of the input correlation matrix, after fitting $n = 50$ samples for $T = 50$ iterations with initialization variance $\s_w^2 = 0.1$ and variance of the ground-truth weights being $\s_*^2 = 2$, {this experiment uses $N=100$ random initializations.}
}
\label{fig:decomposition}
\end{figure}

It might seem counter-intuitive that sloppier data, i.e., large $c$, leads to a slower decay in the eigenspectrum of $P_1^y$. But notice sloppier data will also lead to a faster decay in the tail of the eigenspectrum of $P_1^{\s_w}$. Specifically, in~\cref{fig:w1_eval_bound} the threshold upon the index $i$ after which the eigenspectrum of $P_1^{\s_w}(T)$ decreases quickly is $i^* < \ln(2T \a)/c$. This threshold scales as $1/c$. Therefore, if one trains for small times $T$, the eigenspectrum of the sum $P_1(T) = P_1^{\s_w}(T) + P_1^y(T)$ still decays after $i^*$, essentially dominated by $P_1^{\s_w}$. In other words, for the same $T$, sloppier the data, smaller the threshold $i^*$ after which the eigenspectrum of $P_1(T)$ decays.

{ We should note that although~\cref{lemma:w0_decay_bound} does show that the eigenspectrum of $P_1^y$ decays, it is a loose upper bound. In our experiments, the decay of the eigenspectrum is typically about twice as fast. This is because our upper bound depends on the displacement rank of the matrix, and the decay rate of the bound for a matrix with a displacement rank of 2 is twice as slow as that of a matrix with a displacement rank of 1. In our case, the displacement rank is the rank of the matrix $yy^\top - K_d^T yy^\top K_d^T$ in $P_1^y(T)$ which is 2, but in our numerical experiments we observe the decay rate to be closer to the bound with displacement rank of 1. This suggests that there could be better ways to exploit the displacement structure and improve our bounds. Note that if the labels $y_i$ are aligned with an eigendirection of the data then  $[yy^\top,K]=0$, and the rank of the above matrix is 1. In this case the training manifold geometry can be calculated analytically and is dominated by the decay of the initialization given in \cref{eq:lambda_sw}. }   

\medskip \noindent \textbf{Combining the two parts to obtain the eigenspectrum of $\mathbf{P_1(T)}$.}
The following lemma combines the technical development in the previous two subsections.

\begin{lemma}
\label{lemma:linear_gd_pca_decay_rate}
The eigenvalues of $P_1(T)$ in~\cref{eq:P} satisfy
\[
\aed{
\frac{\l_i^{P_1}}{\l_1^{P_1}}
&\leq  \min \cbr{1, 4\rho^{-(i-1)} + \frac{\s_w^2}{\a \norm{y}^2}} & \text{if } i \leq 2 k^*,\\
&\leq 4 \rho^{-(k^*-1)} + \frac{\s_w^2}{\a \norm{y}^2}  \min\{1, 2 \a T \l_{i-k^*+1}^K\} & \text{else}.
}
\]
for all $i=1,\dots, n$, where $k^* = \min \cbr{\f{\ln 2T\a}{2c}, \f{n}{2} }$.
\end{lemma}

\begin{figure}[!htpb]
\centering
\includegraphics[width=0.65\linewidth]{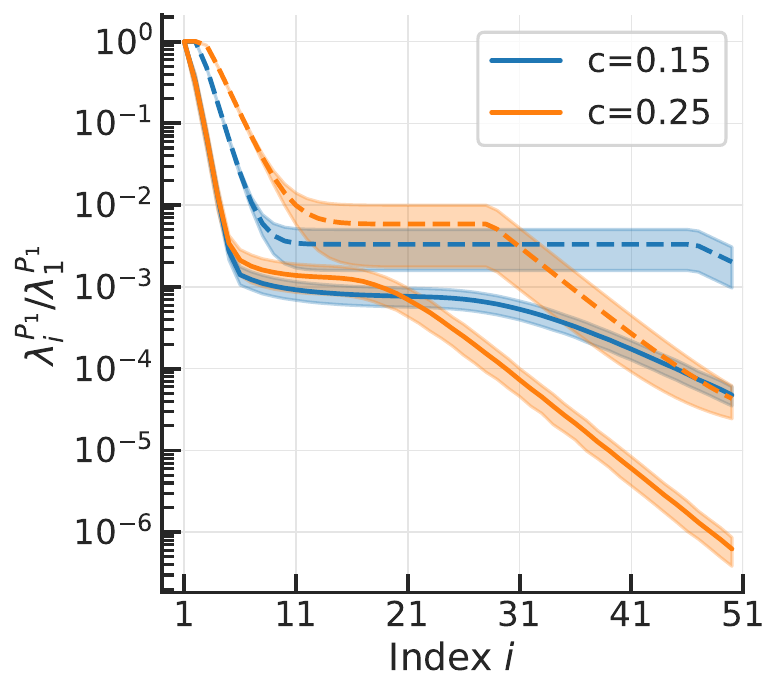}
\caption{
\textbf{Comparison of the bound in~\cref{lemma:linear_gd_pca_decay_rate}} (dashed)
\textbf{with eigenvalues of $P_1(T)$ computed directly from~\cref{eq:P}}
(solid) \textbf{for different values of sloppy decay rate $c$.} { We use 100 random initializations for each experiment, error bars in the above plot denote standard deviation across 100 numerical experiments.}
The proof of~\cref{lemma:linear_gd_pca_decay_rate} works by computing the ideal point to apply Weyl's inequality. This enables us to separately calculate the decay in the head of the eigenspectrum and the tail, for both small and large training times $T$ in spite of the fact that different parts of $P_1(T)$ in~\cref{eq:P} dominate in different regimes.
}
\label{fig:bound_vs_pt}
\end{figure}

\begin{figure}[!htpb]
\centering
\includegraphics[width=0.65\linewidth]{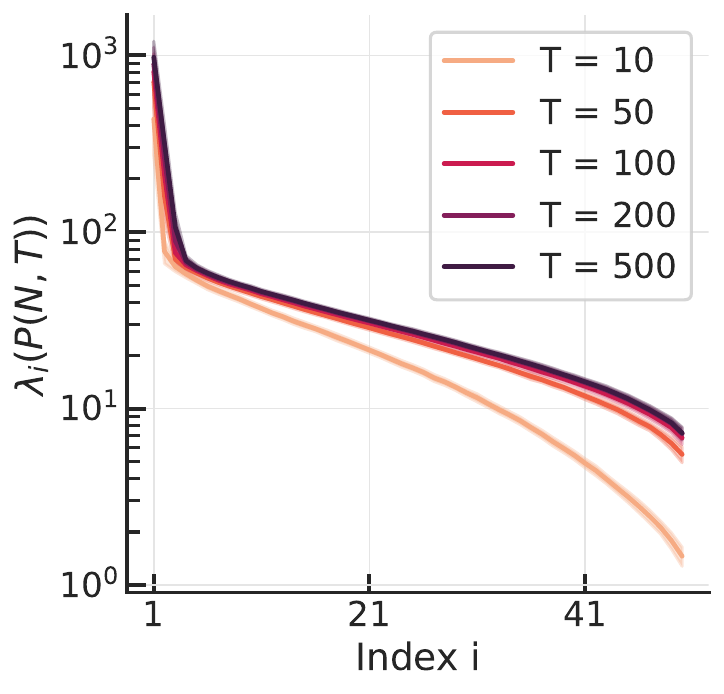}
\caption{\textbf{Eigenvalues of $P(N,T)$ for different training times $T$ from numerical experiments on linear models} with $d=100$ dimensional data with $n=50$ training samples, slope $c=0.1$, initialization variance $\s_w^2 = 1$ and learning rate $\a = 1/\l_1^K$. { We use 100 random initializations for each experiment, error bars in the above plot denote standard deviation across 100 numerical experiments.}
As training time $T$ increases, the eigenvalues in the tail increase in magnitude, this is because $P_1^{\s_w}(T)$. See~\cref{fig:w1_eval_bound}.
This also causes an increase in the largest eigenvalue in the head, due to the diminishing magnitude of $P_2(T)$ in~\cref{eq:P2}.
}
\label{fig:finite_time_evals}
\end{figure}

\cref{fig:bound_vs_pt} compares the upper-bound on the eigenvalues derived from \cref{lemma:linear_gd_pca_decay_rate} against the eigenvalues computed directly from~\cref{eq:P} for different values of sloppy decay rate $c$. This lemma correctly predicts the qualitative trends in the eigenspectrum, in the head of the spectrum where it decays quickly, in the intermediate plateau where eigenvalues do not decay, and in the tail where it again decays quickly. \cref{fig:finite_time_evals} shows, using a numerical calculation, how the eigenspectrum of $P(N, T)$ in~\cref{eq:rpca_finite} changes for different training times $T$.

\begin{figure*}[!htpb]
\centering
\subfloat[$\s_*/\s_w = 4.38$]
{\includegraphics[width=0.31\linewidth]{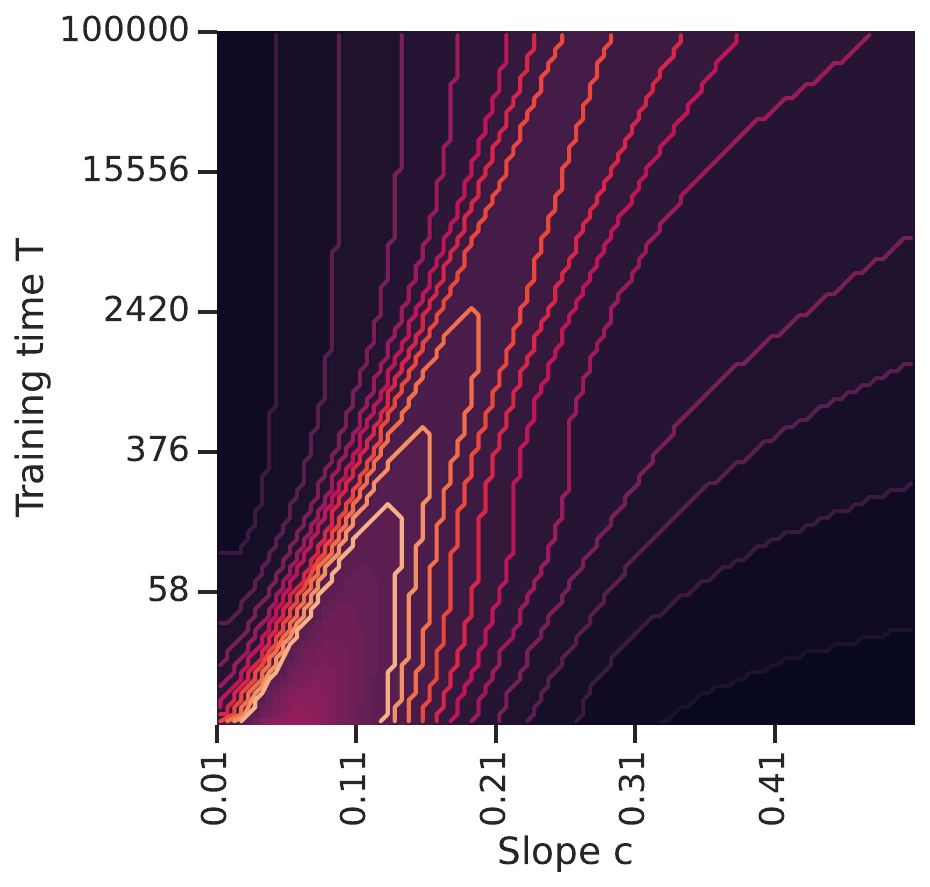}}
\label{fig:sigma_1}
\hspace*{1em}
\subfloat[$\s_*/\s_w = 1.32$]
{\includegraphics[width=0.25\linewidth]{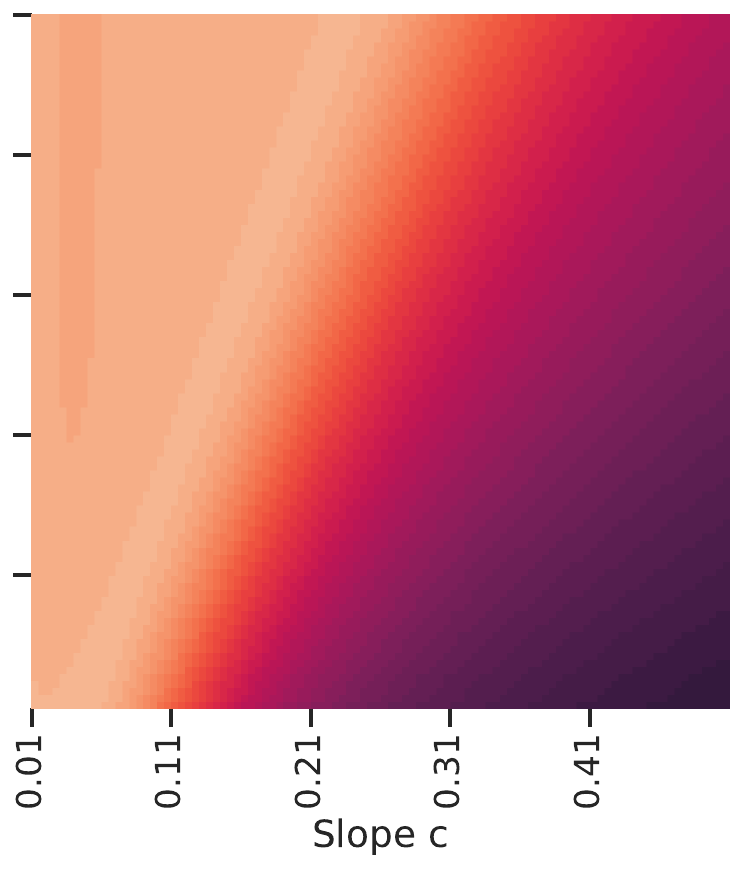}}
\label{fig:sub2}
\hspace*{1em}
\subfloat[$\s_*/\s_w = 0.33$]
{\includegraphics[width=0.3\linewidth]{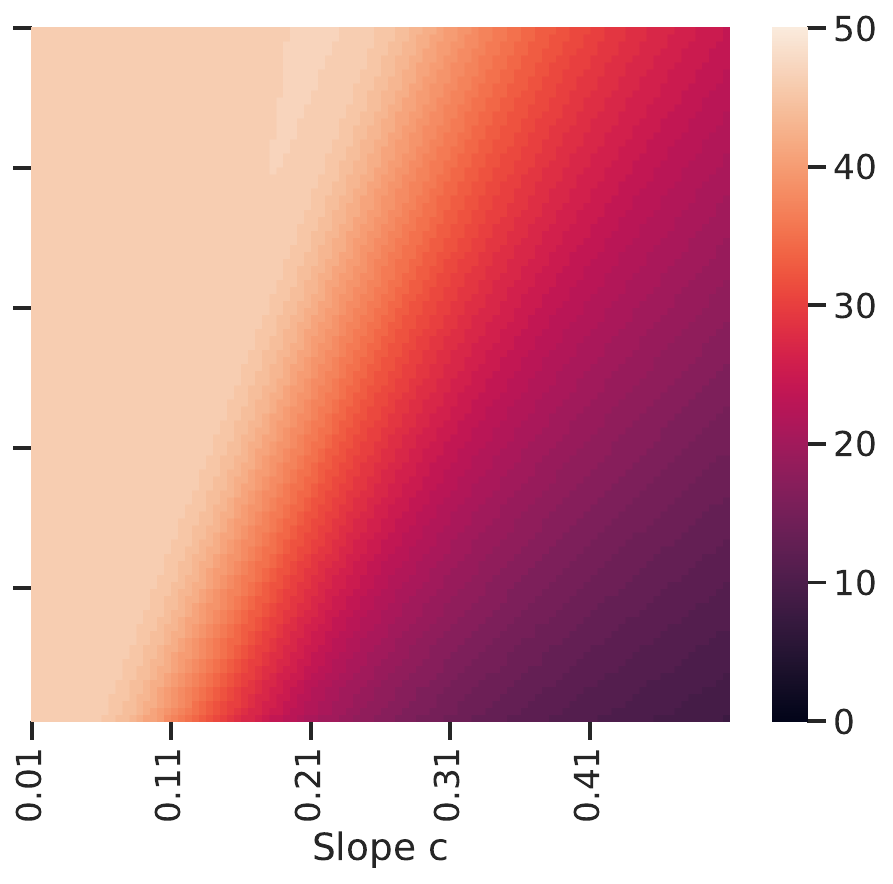}}
\label{fig:sub3}
\subfloat[Transitions between phases of the hyper-ribbon]
{\includegraphics[width=0.4\linewidth]{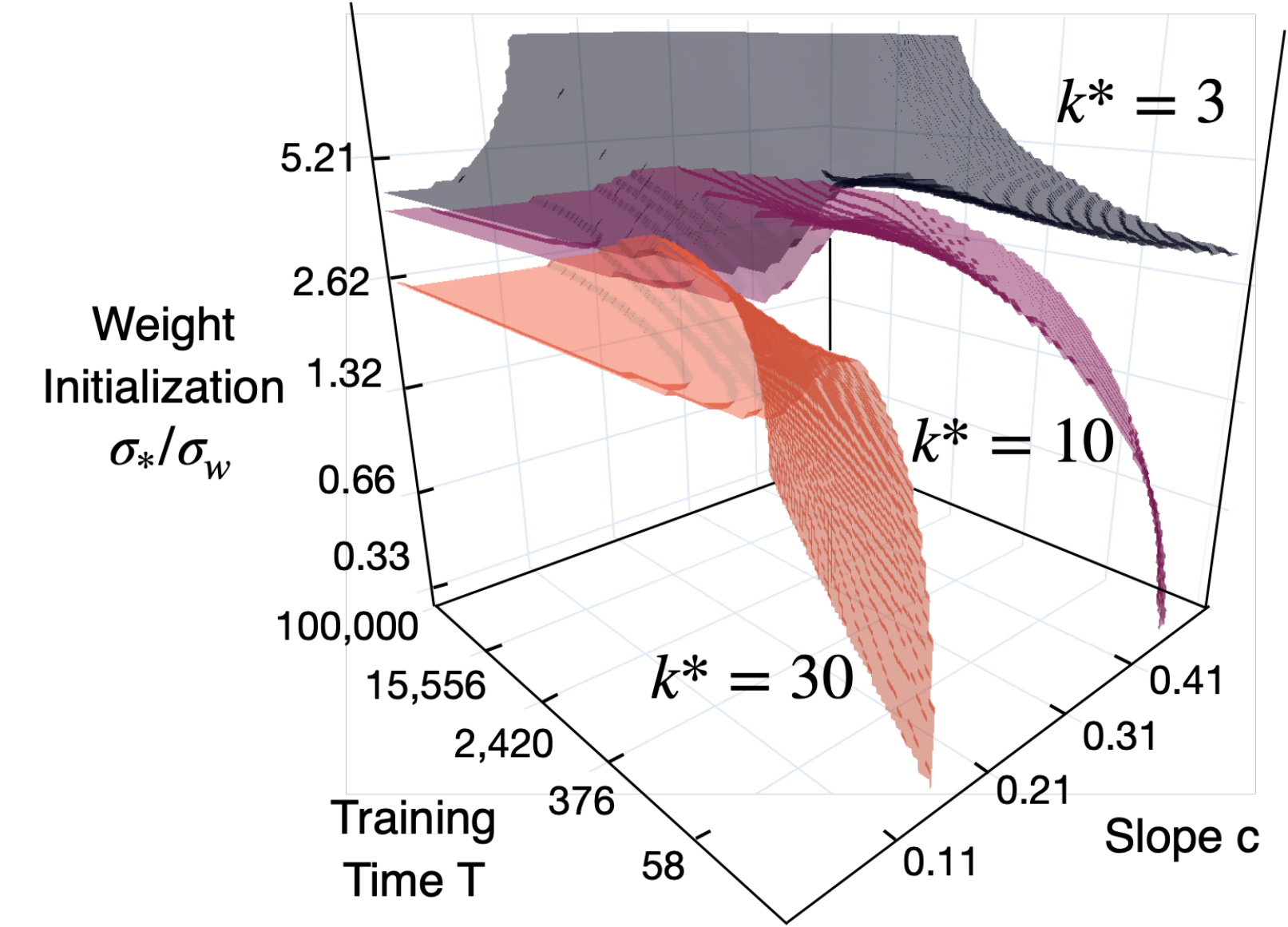}}
\label{fig:3d}
\subfloat[$k^* = 3$ dimensions]
{\includegraphics[width=0.31\linewidth]{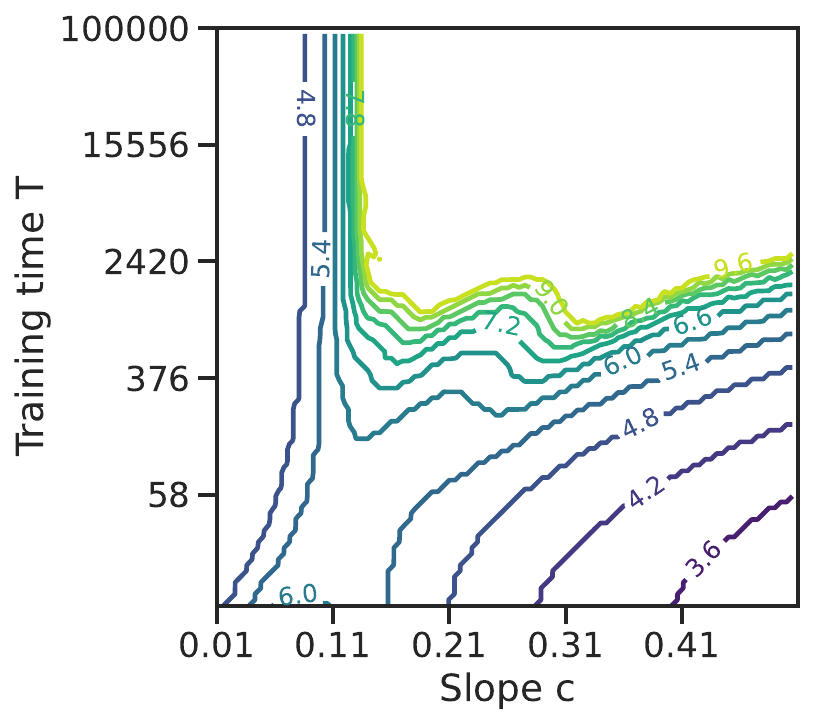}}
\label{fig:contour_3dims}
\subfloat[$k^* = 10$ dimensions]
{\includegraphics[width=0.25\linewidth]{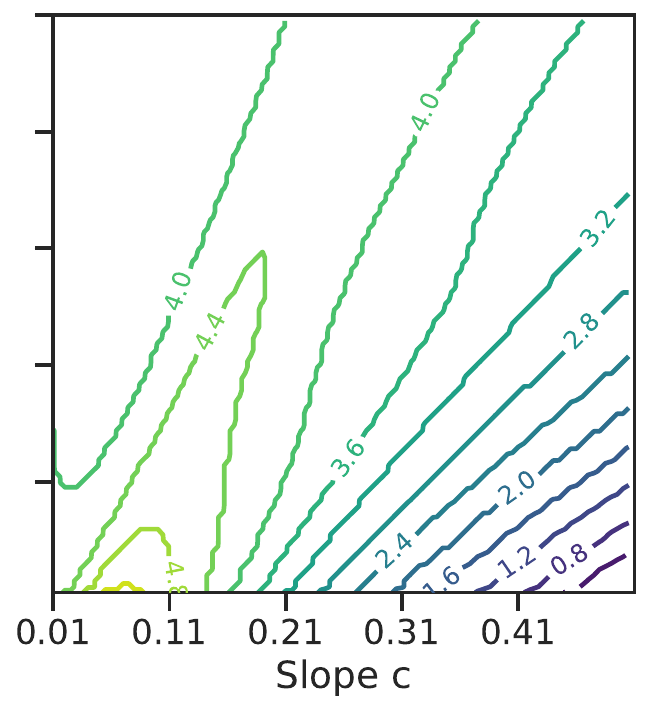}}
\label{fig:contour_10dims}
\caption{
\textbf{A phase diagram for linear regression that describes the number of dimensions in the hyper-ribbon, i.e., the number of dimensions required to capture 95\% of the variance of the points on the training manifold.}
This is studied with respect to three parameters: (i) the training time $T$, (ii) slope $c$, and (iii) the relative magnitude of weight initialization $\s_*/\s_w$.
\textbf{(a-c)} show a heat-map of the dimensionality of the hyper-ribbon for different training times $T$ and slopes $c$ for three different regimes of weight initialization.
\textbf{(d)} is a three dimensional plot that depicts the boundaries of the different phases, defined by the dimensionality of the hyper-ribbon (3 dimensions in black, 10 in pink and 30 in orange).
\textbf{(e-f)}
show contours for different values of $\s_*/\s_w$ for three and ten- dimensional hyper-ribbons, respectively.
See the narrative for an elaborate discussion.
}
\label{fig:contour_plots}
\end{figure*}

\medskip \noindent \textbf{Phase diagrams.}
\cref{fig:contour_plots} summarizes the development of this section using a phase diagram that describes the geometry of the hyper-ribbon  in terms of the relevant parameters, the training time $T$, the slope $c$ and the relative magnitude of the weight initialization $\s_*/\s_w$.
Consider \cref{fig:contour_plots} (a). For small initialization variance $\s_*/\s_w \gg 1$, the hyper-ribbon is very low-dimensional for most training times $T$ and slope $c$. The eigenspectrum is dominated by $P_1^y$ in this case and its fast decay allows for lower-dimensional hyper-ribbons. There appears to be a straight line ($\log T \propto c$) along which the dimensionality is larger, due to relative magnitudes of $P_1^{\s_w}$ and $P_1^y$ in~\cref{eq:P}. The matrix $P_1^y$ results in a higher-dimensionality for large $c$ while $P_1^{\s_w}$ is the cause of higher-dimensionality at relatively small values of $c$ and large $T$. For small $T$, the relatively large magnitude of $\lambda^{P_2}$ will reduce the initial part of the eigenspectrum (coming mostly from $P_1^y$), resulting in a higher dimensionality.
If the initialization variance is small, short training times do not fit the data well. For large $c$, this causes the hyper-ribbon to be low-dimensional (roughly, because the condition number of optimization is large and different models end up being rather similar). The majority of experiments in~\cite{mao2024training} lie in this regime. For small $c$, this is evident as a higher-dimensional hyper-ribbon (roughly, because models are initialized in different subspaces of the data). For longer training times $T$, different models fit the data very well when $c$ is small (again, because of a benign condition number). This is evident as a low-dimensional hyper-ribbon above the straight line.

Next consider \cref{fig:contour_plots} (b-c). As the initialization variance increases, the apparent straight line $\log T \propto c$ that distinguishes low-dimensional hyper-ribbons from higher-dimensional ones, is still present. The upper-left region is increasingly higher-dimensional. For small slope $c$ the hyper-ribbon is high-dimensional for all training times $T$. Because, models are initialized in very different subspaces of the data, and this is true for all three plots, except that it becomes more apparent as $\s_*/\s_w$ decreases. For large $c$, for small times, the hyper-ribbon may be low-dimensional but we need much longer times to fit this data well.
\cite[Fig. 10, S.10, S.16]{mao2024training} showed that when neural networks are initialized very far away from standard initializations, the hyper-ribbon is not low-dimensional. The dimensionality further increases when input data is not sloppy. Their experiments lie in regimes \cref{fig:contour_plots} (b-c).

{
The training dynamics of deep linear networks has an initial ``alignment'' phase where weights rotate from their initial values towards the eventual solution, and a second phase where the dynamics predominantly changes the magnitude of the weights \cite{atanasovNeuralNetworksKernel2021,shamir2021dimpled}. Linear networks considered here do not have the initial alignment phase, so our results in \cref{fig:sigma_1} (a) hold for the deep linear networks in the second phase. In other words, we conjecture that the dimensionality of the training manifold in \cref{fig:sigma_1} (a) is a lower-bound for the dimensionality of the manifold of an equivalent deep linear network. The contribution to the dimensionality coming from the initial alignment phase diminishes as the number of weight updates $T$ increases.
}

Next consider \cref{fig:contour_plots} (d). A three dimensional plot that depicts the boundaries of the different phases, defined by the dimensionality of the hyper-ribbon (3 dimensions in black, 10 in pink and 30 in orange). Some broad trends are apparent in the 3D plot, e.g., (i) large $\s_*/\s_w$ leads to a low-dimensional hyper-ribbon, (ii) the geometry of the hyper-ribbon is very sensitive to other parameters when the slope $c$ is small. As one goes from small $T$, large slope $c$ and large $\s_*/\s_w$, to large times, small slope and small $\s_*/\s_w$, the dimensionality of the hyper-ribbon increases. The other panels in this figure are obtained by projecting this phase diagram upon different axes.

\cref{fig:contour_plots} (e-f)
show contours for different values of $\s_*/\s_w$ for three and ten- dimensional hyper-ribbons, respectively. Focus on the contour marked $4.8$, the two left and right wings of this contour together lead to a slice of \cref{fig:contour_plots} (a) at a fixed dimension of three. \cref{fig:contour_plots} (e) indicates that there is a contiguous region in $(T, c, \s_*/\s_w)$-space with $\s_*/\s_w \gtrapprox 9$ where the hyper-ribbon has fewer than three dimensions. In \cref{fig:contour_plots} (f) such contiguous regions occur at small values of $\s_*/\s_w$.

Altogether, \cref{fig:contour_plots} (d-f) shed light on thumb-rules for identifying the complexity of models that would be required to fit data in these different regimes. Given a dataset (a proxy for its complexity would be $c$), a training recipe (a proxy of which would be $\s_*/\s_w$) and training budget (a proxy of which would be $T$), the boundary of the phase diagram indicates the smallest model that one needs to achieve a good fit. For example, if our regime lies below the orange surface, we need to fit a model with a larger number of parameters.

{
Sloppy models such as deep networks exhibit spectral bias where the training dynamics fits to the top eigen-components of the data at early times \cite{saxeExactSolutionsNonlinear2014, Bartlett_Long_Lugosi_Tsigler_2020, Bordelon_Canatar_Pehlevan_2020}.
Low-dimensional training manifolds, identified experimentally in \cite{mao2024training} and analytically in this paper, are not directly related to this phenomenon. The training manifold is low-dimensional due to the similarity of (the predictions of models along) trajectories initialized at different locations, e.g., when networks fit similar subspaces, not necessarily the top eigenspace, at similar rates. Low-dimensional training manifolds exist even when the network fits to the data well, e.g., in \cref{fig:sigma_1}(a), the dimensionality is small for very large times, especially when $c \ll 1$. Spectral bias does imply that the training manifold is low-dimensional for a single architecture. But since different architectures have different spectral biases \cite{choraria2022spectral}, this does not capture the low-dimensionality of the training manifolds for diverse architectures that was reported experimentally, especially in view of \cite[Fig. 11]{mao2024training} and \cite{hacohen2020let} which showed that different architectures, optimization and regularization-based configurations fit the same easy images in the dataset first and the same challenging images towards the end of training.
}

\section{Variants of the linear model}
\label{s:modifications}

{
We next extend our analysis to (i) stochastic optimization algorithms, (ii) $\ell_2$ regularization of the weights, and (iii) kernel machines. In all three cases, we will see that the PCA matrix $P(N,T)$ in \cref{eq:rpca_finite} undergoes minor changes and can be analyzed using \cref{lemma:linear_gd_pca_decay_rate} to show that the training manifold is low-dimensional. This section sheds further light into explaining the experimental results of \cite{mao2024training}---just like we showed in the prequel that the manifold of networks with the same architecture (linear) trained with gradient descent from different initializations is low-dimensional, the manifolds obtained from different optimization algorithms, regularization techniques and architectures are also low-dimensional. In each of the three settings, we will also be able to interpret the decay of the eigenvalues of $P(N,T)$ to provide thumb-rules for hyper-parameter selection.
}

\medskip \noindent \textbf{Stochastic Gradient Descent (SGD).}
Let us now consider stochastic gradient descent for the linear predictive model. In this case, the weights $w \in \R^d$ are updated, not using the gradient on the entire objective as before $w_{t+1} = w_t - \a \partial_w C(w_t)$, but instead as
\[
    w_{t+1} = w_t - \f{\a}{2 \bb} \partial_w \cbr{\sum_{i=1}^{\bb} r^2_{\w_i}(w_t)}.
\]
where the random variable $\w_i$ is uniformly distributed on $\{1,\dots,n\}$ and $\bb$ is the batch-size. We can model SGD as gradient descent perturbed by state-dependent Langevin noise
\beq{
    w_{t+1} = w_t - \a \partial_w C(w_t) + (\a/\sqrt{\bb}) \xi_t,
    \label{eq:sgd}
}
where $\xi_t \sim N(0, D)$ where $D = X^\top X/n - \bar x^\top \bar x$ with $\bar x = n^{-1} \sum_{i=1}^n x_i$ is the covariance matrix of the inputs~\citep{chaudhari2017stochastic}. Under this dynamics of the weights, the residuals evolve as
\[
r_{t+1} = (I - \a K) r_t + (\a/\sqrt{\bb}) X \xi_t.
\]
The PCA matrix $P(N, T)$, as the number of random initializations $N$ goes to infinity, becomes
\[
P(T) = \frac{1}{T} \sum_{t = 0}^{T-1} \E \sbr{r_t r_t^\top} - K_T y y^\top K_T,
\]
where $K_T$ is the same matrix as in~\cref{eq:P}.
Notice here the randomness of $r_t$ comes from both random initialization and noise from Langevin dynamics, and we are taking expectation with both sources of randomness, assuming they are independent. We now have
\begin{align*}
&\E \sbr{r_{t+1} r_{t+1}^{\top}}\\
&= K_d \E  \sbr{r_t r_t^{\top}} K_d^{\top}
+ 2 (\a/\sqrt{\bb}) K_d \E \sbr{r_t} \E \sbr{\xi_t} X^{\top}\\
&\qquad + (\a^2/\bb) X \E \sbr{\xi_t \xi_t^{\top}} X^{\top}\\
&=  K_d \E  \sbr{r_t r_t^\top} K_d^\top + (\a^2/\bb)  X X^\top X X^\top \\
&= K_d \E \sbr{r_t r_t^\top} K_d + (\a^2/\bb) K^2,
\end{align*}
where we recall that $K_d = I - \a K$.

If we define $P_\xi = (\a^2/\bb) \sum_{t = 0}^{\infty} K_d^t K^2 K_d^t$, then $P_\xi$ solves the Lyapunov equation $K_d P_\xi K_d^\top - P_\xi + (\a^2/\bb) K^2 = 0$, and it can be checked by induction that
\begin{align*}
&\sum_{t = 0}^{T-1} \E \sbr{r_t r_t^\top}
= T P_\xi + \sum_{t = 0}^{T-1} K_d^t \rbr{\E \sbr{r_0 r_0^\top} - P_\xi} K_d^t\\
&= \sum_{t = 0}^{T-1} K_d^t \rbr{\s_w^2 K + yy^\top} K_d^t
+ T P_\xi - \sum_{t = 0}^{T-1} K_d^t P_\xi K_d^t,
\end{align*}
which has two additional terms  compared to~\cref{eq:P}.
Notice that $K_d$ commutes with $K^2$ so $P_\xi$ has the explicit expression
\[
P_\xi = (\a^2/\bb) (I - K_d^2)^{-1} K^2 = (\a/\bb) (2I-\a K)^{-1}K.
\]
Since $P_\xi$ commutes with $K_d$, we can simplify
\begin{align*}
\sum_{t = 0}^{T-1} K_d^t P_\xi K_d^t
&= (\a/\bb) \sum_{t = 0}^{T-1} K_d^{2t} (2I-\a K)^{-1} K \\
&= (1/\bb) (2I - \a K)^{-2} (I - K_d^{2T})
\end{align*}
For $\a \l_i^K < 1$ the $i^{\text{th}}$ eigenvalue of the above matrix is $\simeq T\l_i^K / (2 \bb)$.
In~\cref{lemma:linear_gd_pca_decay_rate}, the upper bound on the eigenvalues of $P_1(T)$ for $i \leq 2k^*$ is perturbed by largest eigenvalue of the two additional terms. This is at most the largest eigenvalue of $P_\xi$, which is simply $\l_1(P_\xi) = (\a/\bb) (\l_1^K/(2 - \a \l_1^K)) \leq \a/\bb$ for our setting where $\l_1^K = 1$ and $\a < 1$ (which ensures the convergence of the infinite sum in $P_\xi$). In other words in~\cref{lemma:linear_gd_pca_decay_rate}, we will have,
\[
\frac{\l_i^{P_1}}{\l_1^{P_1}}
\leq  \min \cbr{1, 4\rho^{-(i-1)} + \frac{1}{\norm{y}^2} \rbr{ \f{\s_w^2}{\a} + \f{\a}{\bb} } }
\]
for $i \leq 2k^*$.

This indicates an interesting relationship between the variance of weight initialization $\s_w^2$, the learning rate $\a$ and the batch-size $\bb$. Suppose we wish to keep the volume of the ensemble of trajectories, as measured by the volume of the hyper-ribbon in residual space, the same. This is a reasonable because it indicates the propensity to identify good fits within the ensemble. Suppose we are in the regime where $\a \propto \bb$, which is very common while training neural networks. If we pick a batch-size that is twice as large, the first term $\s_w^2/\a$ shrinks by a factor of two. To compensate---to keep the decay rate and effectively the dimensionality of the hyper-ribbon the same---we must pick a weight initialization variance that is twice as large.

\medskip \noindent \textbf{Weight Decay}
The least squares objective~\cref{eq:C} is often ``regularized'' to be $C(w) + \f{\l}{2} \norm{w}_2^2$ to obtain a fit where the weights have a small $\ell_2$-norm. This is important in the data scarce regime, i.e., $n < d$, where there may be multiple solutions to the non-regularized problem. The residual dynamics can be seen to be $r_{t+1} = (I - \a K_\l) r_t$ where $K_\l = X X^\top + \l I_{n \times n}$. All our calculations in~\cref{s:linear_regression} hold with $K$ replaced by $K_\l$, i.e., with each $\l_i^{K_\l} = \l_i^K + \l$. For example, from~\cref{lemma:linear_gd_pca_decay_rate}, for $i \leq 2k^*$ we have
\[
\frac{\l_i^{P_1}}{\l_1^{P_1}}
\leq  \min \cbr{1, 4\rho^{-(i-1)} + \frac{1}{\norm{y}^2} \rbr{ \f{\s_w^2}{\a} + \l} }
\]
which indicates that the decay rate of the eigenvalues is now a balance between the regularization parameter $\l$ and the ratio $\s_w^2/\a$.

\medskip \noindent \textbf{Kernel machines.}
Consider predictions given by $\hat y_i = f(x_i)$ where the predictor $f$ is not linear, but it belongs to a class of functions $\FF$ with some regularity properties. A classical example of such a class of functions is called reproducing kernel Hilbert space (RKHS) which is a Hilbert space with the ``reproducing kernel property''. This property states that for any input datum $x$, there exists a function $\varphi_x \in \FF$ such that the evaluation of $f \in \FF$ at the input $x$ can be written as an inner product $f(x) = \inner{f}{\varphi_x}_\FF = \int f(x') \varphi_x(x') \dd{x'}$. The function
\[
k(x,x') = \inner{\varphi_x}{\varphi_{x'}}_\FF \geq 0
\]
is called the reproducing kernel. Gradient descent in RKHS~\cite{Yao_Rosasco_Caponnetto_2007} to minimize the objective in~\cref{eq:C} corresponds to updates of the form
\[
    \forall x:\ f_{t+1}(x) = f_t(x) - \f{\a}{n} \sum_{i=1}^n \rbr{f_t(x_i) - y_i} k(x_i, x).
\]
Notice that the residuals $r_i = f_t(x_i) - y_i$ for all $i=1,\dots,n$ follow linear dynamics, same as those in~\cref{eq:rtp}, namely,
\(
    r_{t+1} = (I - \a K) r_t
\)
except that we now have $K_{ij} = k(x_i, x_j)/n$ for any $i,j$. This matrix is called the Gram matrix and it is positive semi-definite by Mercer's theorem~\cite{scholkopf2002learning}. In other words, all our development in the previous section holds for trajectories of kernel machines initialized from different initial conditions.

If the input correlation matrix is sloppy, then the Gram matrix is sloppy.
This is easiest to see if $x$ comes from a high dimensional distribution, i.e., $d \to \infty$ as $n \to \infty$ with a fixed $d/n$. Results in random matrix theory~\cite{Karoui_2010} state that the Gram matrix $K$ can be well-approximated by the sample covariance matrix in such cases. And therefore, kernel machines are rather similar to linear models. If inputs $x$ are drawn from a distribution with density $p(x)$ supported on $\R^d$, as $n \to \infty$, the $k$-th eigenvalue of the Gram matrix $K$ converges to the $k$-the eigenvalue of the integral operator $T[\varphi](x) = \int k(x, x') \varphi(x') p(x') \dd{x'}$~\cite{Koltchinskii_Giné_2000}. The eigenspectrum of such integral operators has been studied, e.g., if $p(x)$ is Gaussian and $k(x,x') \propto \exp(-\norm{x-x'}^2/d)$ is highly local, then eigenvalues of the Gram matrix $K$ decay exponentially~\cite{Zhu_Williams_Rohwer_Morciniec}. For translation invariant kernels, the decay is related to how quickly $p(x)$ goes to zero with increasing $\norm{x}$ and to the Fourier transform of the kernel $k$~\cite{Widom_1963}. In other words, the Gram matrix $K$ can also be sloppy even if the input correlation matrix is not.


Let us now consider the space of training trajectories corresponding to different kernel models. Let $\{K^{(m)}\}_{m=1}^M$ be $M$ different kernel machines with trajectories in the residual space defined by $r_{t+1}^{(i,m)} = (I - \a K^{(m)})^{t+1} r_0^{(i,m)}$ for the $i^{\text{th}}$ initialization. This suggests that we should study the covariance matrix
\[
    P(N, M, T) = \frac{1}{NMT} \sum_{i, m, t} \rbr{r_t^{(i,m)} - \bar r} \rbr{r_t^{(i,m)} - \bar r }^\top,
\]
where the mean $\bar r = \frac{1}{MNT} \sum_{i, m, t} r_t^{(i, m)}$. Taking $N \to \infty$ as before, we get
\[
P(M, T) \equiv \frac{1}{M} \sum_{m = 1}^{M} P_1^{(m)}(T) - P_2(M, T),
\]
where $P_2(M, T) = K_{T, M} y y^\top K_{T, M}$ with
\[
K_{T, M} = \frac{1}{M T} \sum_{m = 1}^{M} \sum_{t = 0}^{T-1} {K^{(m)}_d}^t.
\]
The above equation is the analog of~\cref{eq:P}. Similar to our previous analysis, $P_2(M, T)$ has a single eigenvalue that vanishes for large $T$.
We can thus obtain a result that is rather similar to the one in~\cref{lemma:linear_gd_pca_decay_rate} with appropriate substitutions $\r \to \r^{(m)}$ and $\l_i^K \to \l_i^{K^{(m)}}$, assuming different kernels have different slopes $c^{(m)}$ but the same largest eigenvalue, i.e., the same learning rate $\a$.
Now by Weyl's inequality,
\[
\lambda_{M(i-1)+1} \left( \sum_{m=1}^M P_1^{(m)}(T) \right) \leq \sum_{m=1}^M \lambda_i\left( P_1^{(m)} (T) \right)
\]
and the fact that $\lambda_1(\sum_m P_1^{m}) \geq M \norm{y}^2$, the right-hand side of the per-kernel version of~\cref{lemma:linear_gd_pca_decay_rate} can be summed up over $m$.
Altogether, the eigenvalues of $T P_1(M,T) \equiv T P_1$ satisfy
\beq{
\frac{\l_i^{P_1}}{\l_1^{P_1}}
\leq  \min \Big\{ 1,
\f{4}{M}\sum_m {\rho^{(m)}}^{-\lfloor\f{i-1}{M} \rfloor} + \frac{\s_w^2}{\a \norm{y}^2} \Big\}
\label{eq:P1_kernel}
}
if $i \leq 2k^*$ and otherwise we have
\[
    \frac{\l_i^{P_1}}{\l_1^{P_1}} \leq \f{1}{M} \cbr{\sum_m 4 {\rho^{(m)}}^{-(k^*-1)} + \frac{\s_w^2 \min\{1, 2 \a T \l_{\lfloor\f{i-1}{M} \rfloor -k^*+2}^{K^{(m)}}\}}{\a \norm{y}^2}  }
\]
for all $i=1,\dots, n$, where $k^* = \min_m \ln(2T\a)/ (2c^{(m)})$. This is a loose upper bound, because it uses the floor $\lfloor\f{i-1}{M} \rfloor$ in the exponent of $\r^{(m)}$. But due to the averaging over $m$, if different kernels have similar condition numbers, i.e., similar $\r^{(m)}$, then the decay rate of eigenvalues $\l_i^{P_1}$ is shallower by a factor of $M$. But they do decay, and we should expect the hyper-ribbon to be low-dimensional. For the second expression when $i > 2k^*$, the second term (coming from $P_1^{\s_w})$ dominates the eigenspectrum. It has become worse due to the presence of $\l_{\lfloor\f{i-1}{M} \rfloor -k^*+2}^{K^{(m)}}$. But we can see that it still indicates a decay in the eigenspectrum at large $i$.

This narrative gives intuition into the experiments of~\cite{mao2024training}  which are discussed in~\cref{s:intro}, where the training manifold for different neural architectures was computed to find that the explained variance of the top few dimensions was quite high. Our discussion suggests that this can arise only if the ``effective kernels'' of all those networks have Gram matrices that decay quickly. If some of them do not decay quickly, then the explained variance would be low.

\section{Discussion}
\label{s:discussion}

\textbf{Contextualization in terms of the broader literature on sloppy models} The central contribution of this paper is a technical result on characterizing the geometry of the manifold of predictions of linear models as they train on different types of data, for different durations, and using different types of architectures and training methods. We argued that there are broad similarities in the training trajectories of linear models and deep networks, not in the weight space where there are vast differences, but in the prediction space. Just like deep networks evolve on low-dimensional manifolds during training, linear models also evolve on low-dimensional manifolds. In the latter case however, tools in dynamics systems theory allow us to characterize the geometry very precisely. We showed that hyper-ribbon-like training manifolds in linear models are controlled by three factors: (i) sloppiness of the input data, (ii) strength of the weight initialization, and (iii) the number of gradient updates.

It informs our growing understanding of sloppy models and information geometry for the analysis of multi-parameter models. Sloppiness is described better, not as the geometry of the set of predictions of the model for different values of its parameters, but rather as the restriction to the set of possible predictions on new data after fitting the model. The former is a property of the functional form of the model and the statistics of the input data. But the latter is also a property of the task---the questions we ask of the model. When the task depends on collective behavior of the system, rather than individual parameters, multi-parameter models, including deep networks, are sloppy. For such models, further training on new data only makes small changes to the prediction space. This is yet another example of the emergence of simplicity from complexity in physics~\cite{quinnInformationGeometryMultiparameter2021}.

{
Let us note that we do not need geometric decay in the eigenvalues of the Gram matrix for our arguments in this paper to be valid, this assumption is adopted essentially to enable analytic computations. \cref{fig:sloppy_eigen} shows the eigenspectra for three different real datasets across diverse data modalities. Such eigenvalue plots are often plotted on a log-log scale (which gives the famous power laws \cite{ruderman1997origins,fieldWhatGoalSensory1994}) or in terms of the density of the eigenvalues (in random matrix theory \cite{chaudhari2016entropy,papyan2018full}). These are different ways of studying the same phenomenon, namely the decay in the eigenvalues. The important structure in natural data lies in the head of the eigenspectrum---decay in the head reflects how salient variations in the data diminish in importance. The log-log scale emphasizes trends in the mid/tail of the eigenspectrum and hides the trends in the head of the eigenspectrum. While this may be important for some problems, in the context of our specific problem and we argue machine learning in general, it is important to study the decay in the head of the eigenspectrum more precisely.
}

\textbf{Perspective on generalization in deep learning}
{
Deep networks often have more parameters than the number of training data. Their remarkable real-world performance therefore defies foundational assumptions about the role of model parsimony in generalization. There is a large amount of literature that studies this question, ranging from the ``double descent'' paradigm \cite{belkin2019reconciling,chen2023learning,transtrum2024aliasing},  arguments that relate wide minima to generalization~\cite{hinton1993keeping,chaudhari2016entropy}, implicit regularization towards low-complexity solutions during training~\cite{soudry2018implicit}, analyses of generalization in the kernel regime \cite{canatar2021spectral,NEURIPS2022_08342dc6,amini2022target,jacot2018neural}, and compression in the mutual information between inputs, representations, and outputs \cite{dziugaiteComputingNonvacuousGeneralization2017,yang2021does,xu2017information,achille2018emergence}. Emergent low-dimensional geometries termed ``hyper-ribbons'' have been argued to lead to generalization in sloppy models \cite{waterfallSloppyModelUniversalityClass2006,waterfall2006universality}.
Roughly, hyper-ribbons refer to the shape of the set of all possible predictions on the training samples made by models within a hypothesis class. This is conceptually similar to notions of capacity in statistical learning theory except that it focuses on the geometry of the set, not just the volume.

Together, these diverse viewpoints suggest that low-dimensional structure---whether in the loss landscape, the data, or the training dynamics---plays a central role in enabling generalization in deep learning. And this commonality perhaps hints at a unified theory of statistical generalization.
There is some recent work \cite{yang2025effective} that takes a step towards such a theory. The authors use tools from contraction theory \cite{LohmillerSlotineContraction1998} to show that the generalization gap for a general deep network trained using gradient flow after time $t$ is given by $r(0)^\top G(t) r(0)$ where $r(0)$ is the vector of residuals at initialization and $G(t)$ is a certain ``effective Gram matrix'' that conceptually captures the volume of the hypothesis space explored during training. This result is similar to information-theoretic generalization bounds \cite{neu2021information}.
It can be shown that if the Gram matrix in the present paper $K$ is sloppy (which was also observed by \cite{yang2021does}), then this effective Gram matrix is also sloppy. In the context of the present paper, this result therefore suggests that low-dimensional training manifolds lead to effective generalization. Deep networks are a very large hypothesis class, and therefore effective generalization is difficult without a sufficiently large number of samples. The only way deep networks could generalize as well as they do is if the training process does not explore the entire hypothesis space. Our paper shows that if input data is sloppy and if the variance of the weights at initialization is small, then the training manifold spans a very small subset of the space of predictions---and this is perhaps why generalization is possible.
}

\section{Acknowledgments}
We thank Andrea Liu for her probing question about sloppiness and deep neural networks.
JM and PC were supported by grants from the Office of Naval Research (N00014-22-1-2255) and the National Science Foundation (IIS-2145164, CCF-2212519).
JPS and IG were supported by NSF DMR-2327094. IG was also supported by an Eric and Wendy Schmidt AI in Science Postdoctoral Fellowship, and is grateful to the Statistical Physics of Computation laboratory at EPFL for their hospitality.
YS was supported by grants from the Army Research Office (ARO) and the Air Force Office of Scientific Research (AFOSR).
MKT was supported by NSF DMR-1753357 and ECCS-2223985.

\bibliography{bib/references, bib/pratik}

\appendix*

\section{Proofs}
\label{app:proofs}

\medskip \noindent \textbf{Proof of~\cref{lemma:w0_decay_bound}}.
We will denote $T P_1^y(T)$ as $P$ in this proof for clearer exposition. We know that $P$ solves the discrete algebraic Lyapunov equation
\[
K_d P K_d^\top - P + \rbr{yy^\top - K_d^T yy^\top K_d^T} = 0.
\]
See~\cite[Chapter 4.3.3]{Antoulas_2005}, such a $P$ also solves the continuous Lyapunov equation $\tilde{K_d} P + P \tilde{K_d}^\top + BB^\top = 0$ with
\[
\aed{
\tilde{K_d} &= (K_d+I)^{-1}(K_d-I),\\
B B^\top &= 2 (K_d+I)^{-1} (yy^\top - K_d^T yy^\top K_d^T) (K_d+I)^{-1}.
}
\]
Notice that $\tilde{K_d}$ is normal and has a bounded spectrum:
\[
\sigma(K_d) \subseteq [-b, -a],
\]
with \(0<a<b<\infty\).
Notice that $B B^\top$ has a rank of at most 2. From~\cite[Theorem 2.1 and Corollary 3.2]{Beckermann_Townsend_2016} we can conclude that
\[
\l_{1 + 2i}^P \leq  4\rho^{-2i} \l_1^P,  \text{ with } \rho =  \exp\left( \frac{\pi^2}{2\log(4 b/a )} \right).
\]
Notice that $\l_i^{\tilde{K_d}} = -\frac{\a \l_i^K}{2 - \a\l_i^K}$, so with our assumption (iii) which states that \(\a<1/\l^K_1\), we have
\[
\frac{b}{a} = \frac{\l_1^K}{\l_n^K} \frac{(2 - \a \l_n^K)}{(2 - \a \l_1^K)} \leq \frac{2 \l_1^K}{\l_n^K} - 1.
\]

\medskip \noindent \textbf{Proof of~\cref{lemma:lambda_y_bound}.}
The lower bound is obtained by seeing that
\[
\l_1^y \geq \max_t \l_1(K_d^t yy^\top K_d^t) \geq \norm{y}^2,
\]
and the upper bound is given by
\[
\l_1^y \leq \sum_t \l_1(K_d^t yy^\top K_d^t) = \tilde{\lambda}_n \norm{y}^2.
\]
Since the true weights $w^*$ are drawn from an isotropic normal distribution with covariance $\s_{*}^2 I$, the norm of the targets concentrates around the trace of the data correlation matrix $\s_{*}^2\sum_{i=1}^n \l_i^K$. Indeed,
\(
\norm{y}^2 = (V^\top w^*)^\top S^2 (V^\top w^*),
\)
where $V$ and $S$, are given by the singular value decomposition (SVD) of the data matrix $X = USV^\top$. Since entries of $V^\top w^*$ are independent random variables, $\norm{y}^2$ is concentrated around the above value by the Hanson-Wright inequality
\[
\P\rbr{\abs{\norm{y}^2 - \s_*^2 \tr(K)} > t} \leq 2 \exp \left[-a \min \left(\frac{t^2}{\s_*^4 \tr(K)}, \frac{t}{\s_*^2}\right)\right]
\]
where $a > 0$ is a constant independent of $t, K$ and $\s_*$.

\medskip \noindent \textbf{Proof of~\cref{lemma:linear_gd_pca_decay_rate}.}
To keep the notation clear, in the proof we will refer to $P_1^ {\s_w}(T)$ and $P_1^y(T)$ as $P^{\s_w}$ and $P^y$ respectively.

Let us first discuss the case for small times $T$. Notice that if $T < \frac{\l_1^K}{2\l_n^K}$, then $2T\a < \a \frac{\l_1^K}{\l_n^K} < e^{c(n-1)} < e^{cn}$. In this case we have $k^* = \frac{\ln(2T\a)}{2c} < \frac{n}{2}$, so we can obtain the following upper bound on $\l_i^{P_1}$
by Weyl's inequality:
\[
\l_i^{P_1} \leq \begin{cases}
\l_i^y + \l_1^{\s_w} & \text{ for } i \leq 2 k^* \\
\l_{k^*}^y + \l_{i - k^* + 1}^{\s_w} & \text{ for } i > 2k^*.
\end{cases}
\]
For $i \leq 2 k^*$, we have
\begin{align*}
\frac{\l_i^{P_1}}{\l_1^{P_1}} \leq \frac{\l_i^{P_1}}{\l_1^y}
\leq 4 \rho^{-(i-1)} + \frac{\l_1^{\s_w}}{\l_1^y}
\leq 4\rho^{-(i-1)} + \frac{\s_w^2 \min \{1, 2 T \a \l_1^K\} }{\a \norm{y}^2}
\end{align*}
where we have used  $\l_1^{P_1} \geq \l_1^y$ since $P^{\s_w}$ is positive definite.
Note that under assumptions (i-ii), $\l_1^K=1$ and $\a < 1$, therefore we do not need the minimum in the second term.
For $i > 2k^*$, we have
\begin{align*}
\frac{\l_i^{P_1}}{\l_1^{P_1}}
&\leq \frac{\l_{i-k^*+1}^{\s_w}}{\l_1^y} + 4 \rho^{-(k^*-1)}\\
&\leq \frac{\s_w^2  \min \{1, 2 \a T \l_{i-k^*+1}^K\} }{\a \norm{y}^2} + 4 \rho^{-(k^*-1)}
\end{align*}
where we have used~\cref{eq:w1_eval_bound} and assumption (ii).

For large times $T \geq \frac{\l_1^K}{2\l_n^K}$, we can choose the splitting point for Weyl's inequality to be $n/2$. We now have
\begin{align*}
\frac{\l_i^{P_1}}{\l_1^{P_1}}
\leq \frac{\l_{i-n/2+1}^{\s_w}}{\l_1^y} + \frac{\l_{n/2}^y}{\l_1^y}
\leq \frac{\l_{i-n/2+1}^{\s_w}}{\l_1^y} + 4 \rho^{-(n/2-1)}.
\end{align*}
If $T \geq \frac{\l_1^K}{2\a\l_n^K}$, then $2 T \a \l_i^K > 1$ for all $i \geq 1$, so by~\cref{eq:w1_eval_bound}
\[
\l_{i-n/2+1}^{\s_w} \leq \f{\s_w^2}{2\a - \a^2\l_{i-n/2+1}^K}
\]
and
\begin{align*}
\frac{\l_{i-n/2+1}^{\s_w}}{\l_1^y}
\leq \f{\s_w^2}{\a \norm{y}^2} \frac{1}{(2- \a \l_{i-n/2+1}^K)}
\leq \f{\s_w^2}{\a \norm{y}^2}.
\end{align*}
For $\frac{\l_1(K)}{2\l_n(K)} \leq T < \frac{\l_1(K)}{2\a \l_n(K)}$, we may still use the split at $k^* = n/4$ in the above calculation, and take the minimum of the above two bounds for $i > n/2$.

\end{document}